%% file: main.tex
\documentclass[sigconf]{acmart}

\usepackage{mathtools}
\usepackage{lipsum}
\usepackage{graphicx}
\usepackage{caption, subcaption}
\usepackage{url}
\usepackage{amsmath, amsfonts, amsthm, bm}
\usepackage{soul}
\usepackage{tabularx, multirow}
\usepackage{dsfont}
\usepackage[ruled,vlined]{algorithm2e}
\usepackage{booktabs}
\usepackage{xcolor, colortbl}
\usepackage{xspace}
\usepackage{resizegather}
\usepackage{makecell}
\usepackage{CJKutf8}
\usepackage{balance}

\newcolumntype{W}{>{\centering\arraybackslash}m{0.08\linewidth}}
\newcolumntype{X}{>{\raggedleft\arraybackslash}m{0.08\linewidth}}
\newcolumntype{Y}{>{\centering\arraybackslash}m{0.04\linewidth}}
\newcolumntype{Z}{>{\centering\arraybackslash}m{0.07\linewidth}}
\newcolumntype{P}{>{\centering\arraybackslash}m{0.19\linewidth}}
\newcolumntype{Q}{>{\raggedright\arraybackslash}m{0.28\linewidth}}

\AtBeginDocument{%
  \providecommand\BibTeX{{%
    \normalfont B\kern-0.5em{\scshape i\kern-0.25em b}\kern-0.8em\TeX}}}

\copyrightyear{2025}
\acmYear{2025}
\setcopyright{rightsretained}
\acmConference[WSDM '25]{Proceedings of the Eighteenth ACM International Conference on Web Search and Data Mining}{March 10--14, 2025}{Hannover, Germany}
\acmBooktitle{Proceedings of the Eighteenth ACM International Conference on Web Search and Data Mining (WSDM '25), March 10--14, 2025, Hannover, Germany}
\acmDOI{10.1145/3701551.3703500}
\acmISBN{979-8-4007-1329-3/25/03}

\makeatletter
\gdef\@copyrightpermission{
  \begin{minipage}{0.3\columnwidth}
   \href{https://creativecommons.org/licenses/by/4.0/}{\includegraphics[width=0.90\textwidth]{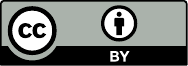}}
   
  \end{minipage}\hfill
  \begin{minipage}{0.7\columnwidth}
   \href{https://creativecommons.org/licenses/by/4.0/}{This work is licensed under a Creative Commons Attribution International 4.0 License.}
  \end{minipage}
  \vspace{5pt}
}
\makeatother

\begin{document}

\title{Unsupervised Robust Cross-Lingual Entity Alignment via~Neighbor~Triple~Matching~with~Entity and Relation Texts}

\author{Soojin Yoon}
\affiliation{
    \institution{Yonsei University}
    \city{Seoul}
    \country{South Korea}
}
\email{soojiny@yonsei.ac.kr}

\author{Sungho Ko}
\affiliation{
    \institution{Yonsei University}
    \city{Seoul}
    \country{South Korea}
}
\email{k133324@yonsei.ac.kr}

\author{Tongyoung Kim}
\affiliation{
    \institution{Yonsei University}
    \city{Seoul}
    \country{South Korea}
}
\email{dykim@yonsei.ac.kr}

\author{SeongKu Kang}
\affiliation{
    \institution{Korea University}
    \city{Seoul}
    \country{South Korea}
}
\email{seongkukang@korea.ac.kr}

\author{Jinyoung Yeo}
\affiliation{
    \institution{Yonsei University}
    \city{Seoul}
    \country{South Korea}
}
\email{jinyeo@yonsei.ac.kr}

\author{Dongha Lee}
\affiliation{
    \institution{Yonsei University}
    \city{Seoul}
    \country{South Korea}
}
\authornote{Corresponding author}
\email{donalee@yonsei.ac.kr}


\begin{abstract}
\input{000abstract}
\end{abstract}

\ccsdesc[500]{Computing methodologies~Semantic networks}
\ccsdesc[500]{Computing methodologies~Ontology engineering}
\ccsdesc[300]{Computing methodologies~Language resources}
\ccsdesc[300]{Computing methodologies~Lexical semantics}

\keywords{Cross-lingual entity alignment; Knowledge graph; Optimal transport; Neighbor triple matching; Pretrained language models}

\input{dfn}



\maketitle

\section{Introduction}
\label{sec:intro}

\input{010introduction}

\section{Related Work}
\label{sec:relatedwork}
\input{020related-work}

\section{Preliminary}
\label{sec:preliminary}
\input{020preliminary}

\section{Method}
\label{sec:method}
\input{030proposed}

\section{Experiments}
\label{sec:exp}
\input{040experiments}

\section{Conclusion}
\label{sec:conclusion}
\input{050conclusion}

\begin{acks}
This work was supported by the IITP grants funded by the Korea government (MSIT) (No. RS-2020-II201361; RS-2024-00457882, National AI Research Lab Project), and the NRF grant funded by the Korea government (MSIT) (No. RS-2023-00244689).
\end{acks}

\pagebreak
\newpage
\clearpage
\bibliographystyle{ACM-Reference-Format}
\balance
\bibliography{bibliography}

\pagebreak
\newpage
\clearpage
\appendix

\input{060appendix}

\end{document}

%% file: 000abstract.tex
Cross-lingual entity alignment (EA) enables the integration of multiple knowledge graphs (KGs) across different languages, 
providing users with seamless access to diverse and comprehensive knowledge.
Existing methods, mostly supervised, face challenges in obtaining labeled entity pairs.
To address this, recent studies have shifted towards self-supervised and unsupervised frameworks.
Despite their effectiveness, these approaches have limitations: 
(1) Relation passing: mainly focusing on the entity while neglecting the semantic information of relations, 
(2) Isomorphic assumption: assuming isomorphism between source and target graphs, which leads to noise and reduced alignment accuracy, and
(3) Noise vulnerability: susceptible to noise in the textual features, especially when encountering inconsistent translations or Out-of-Vocabulary (OOV) problems.
In this paper, we propose \proposed, an unsupervised and robust cross-lingual EA pipeline that jointly performs \ul{E}ntity-level and \ul{R}elation-level \ul{Align}ment by neighbor triple matching strategy using semantic textual features of relations and entities. 
Its refinement step iteratively enhances results by fusing entity-level and relation-level alignments based on neighbor triple matching.
The additional verification step examines the entities' neighbor triples as the linearized text.
This \textit{Align-then-Verify} pipeline rigorously assesses alignment results, achieving near-perfect alignment even in the presence of noisy textual features of entities. 
Our extensive experiments demonstrate that the robustness and general applicability of \proposed improved the accuracy and effectiveness of EA tasks, contributing significantly to knowledge-oriented applications.

%% file: dfn.tex
\renewcommand{\shortauthors}{Soojin Yoon et al.}
\newcommand{\proposed}{\textsc{ERAlign}\xspace}
\newcommand{\proposedpp}{\textsc{ERAlign++}\xspace}
\newcommand{\reranking}{\textsc{Reranking}\xspace}

\newcommand{\seu}{SEU\xspace}
\newcommand{\seupp}{SEU++\xspace}

\newcommand{\eva}{EVA\xspace}
\newcommand{\lightea}{LightEA\xspace}


\newcommand{\veryshortarrow}[1][3pt]{\mathrel{%
\hbox{\rule[\dimexpr\fontdimen22\textfont2-.2pt\relax]{#1}{.4pt}}%
\mkern-4mu\hbox{\usefont{U}{lasy}{m}{n}\symbol{41}}}}
\makeatletter
\definecolor{Gray}{gray}{0.9}

\newcommand{\sinkhorn}{Sinkhorn\xspace}
\newcommand{\lora}{LoRA\xspace}
\newcommand{\iathree}{IA3\xspace}
\newcommand{\finetune}{FineTuning\xspace}
\newcommand{\prompttune}{PromptTuning\xspace}
\newcommand{\prefixtune}{PrefixTuning\xspace}

\newcommand{\tsne}{TSNE\xspace}


\newcommand{\glove}{GloVe\xspace}
\newcommand{\bert}{BERT\xspace}
\newcommand{\sbert}{SBert\xspace}
\newcommand{\roberta}{RoBERTa\xspace}
\newcommand{\robertabase}{RoBERTa\textsubscript{base}\xspace}
\newcommand{\robertalarge}{RoBERTa\textsubscript{large}\xspace}

\newcommand{\dbp}{DBP15K\xspace}
\newcommand{\dbpja}{DBP15K{\scriptsize JA-EN}\xspace}
\newcommand{\dbpzh}{DBP15K{\scriptsize ZH-EN}\xspace}
\newcommand{\dbpfr}{DBP15K{\scriptsize FR-EN}\xspace}

\newcommand{\srp}{SRPRS\xspace}
\newcommand{\srpfr}{SRPRS{\scriptsize EN-FR}\xspace}
\newcommand{\srpde}{SRPRS{\scriptsize EN-DE}\xspace}

\newcommand{\reduce}[1]{\textls[-50]{#1}}
\newcommand{\smallsection}[1]{{\vspace{0.05in} \noindent \bf {#1.\hspace{5pt}}}}

%% file: 010introduction.tex
\begin{CJK*}{UTF8}{gbsn}

\begin{figure}[thbp]
    \centering
    \includegraphics[width=\linewidth]{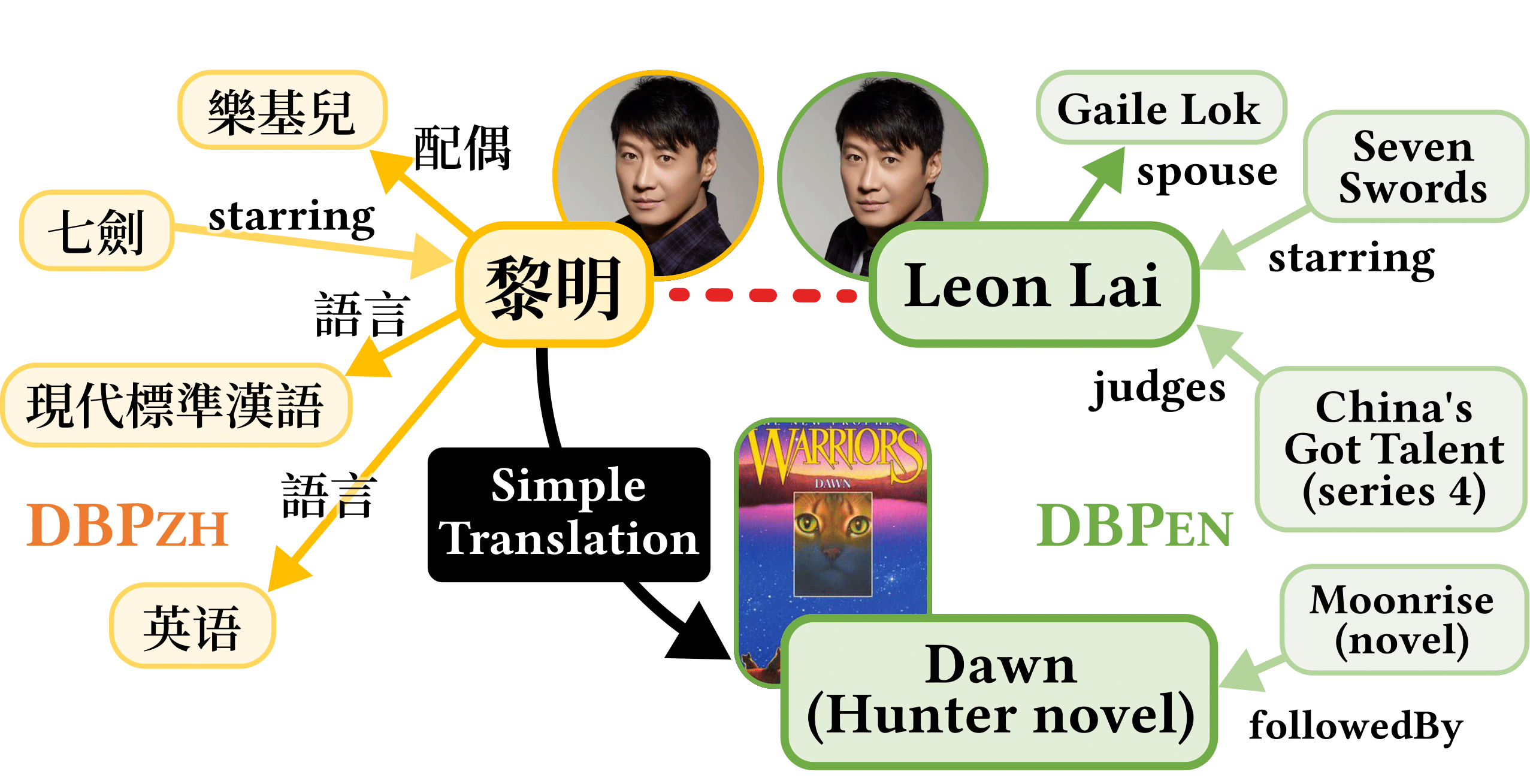}
    \caption{An example of cross-lingual EA on \dbpzh.}
    \label{fig:problem}
\end{figure}

Knowledge graphs (KGs) are structured representations of real-world knowledge~\cite{kgs_dbpdia, kgs_yago, kgs_nell}
and are widely used in applications such as question answering~\cite{qa1,qa2}, commonsense reasoning~\cite{cr1, cr2}, and recommendation~\cite{rec1, rec3}.
Most KGs are developed for specific domains and languages, leading to distinct knowledge bases with overlapping information.
This led to the cross-lingual entity alignment (EA)~\cite{seu, jape}, which identifies equivalent entities across languages.
EA is not trivial, as entities often have proper nouns as names, making simple machine translation unreliable.
Figure~\ref{fig:problem} is an example of the entity ``黎明'', which refers to the Hong Kong actor ``Leon Lai'', but it translates to ``dawn'', which risks misalignment.
Moreover, entities with the same name may represent different concepts, or a single entity may have multiple names, futher complicating alignment.
EA tackles these issues, enabling seamless integration of knowledge across multilingual KGs.
\end{CJK*}

Most cross-lingual EA methods adopt a \textit{supervised} approach, training a {parametric encoder}~\cite{transe, gcn-align, gm-align, srea} with structural feature from translation embeddings~\cite{transe, mtranse, jape, bootea} or Graph Neural Networks (GNNs)~\cite{gcn-align, rdgcn, hgcn}.
However, they rely on pre-aligned pairs, which are often challenging to obtain.
Recent advances include \textit{self-supervised} and \textit{unsupervised} frameworks.
Self-supervised methods generate pseudo-seed pairs from auxiliary data (e.g., images or texts) to train without pre-aligned pairs,
while unsupervised methods treat EA as an Optimal Transport (OT) problem,
using Hungarian~\cite{hungarian} or Sinkhorn~\cite{sinkhorn} algorithms to compute alignments.

Despite effectiveness, these methods have limitations:
(1) \textbf{Relation passing}: They mainly focus on entity features, neglecting the semantic information of relations,
which limits their ability to capture heterogeneous relationships.
(2) \textbf{Isomorphic assumption}: They assume that source and target graphs are isomorphic, a condition rarely met in practice.
(3) \textbf{Noise vulnerability}: They are susceptible to textual noise, such as inconsistent translations or Out-of-Vocabulary (OOV) issues, reducing accuracy.

We propose \proposed, a robust cross-lingual EA pipeline that jointly performs \ul{E}ntity-level and \ul{R}elation-level \ul{Align}ment. 
Unlike existing methods, \proposed leverages semantic textual features of both entities and relations to capture \textit{neighbor triples} for alignment, enhancing robustness and efficacy through triple-level semantic matching.
Using \textit{dual knowledge graphs}, where entities and relations are represented as edges and nodes, respectively,
enabling align relations without additional techniques.
Alignment scores are iteratively refined through \textit{neighbor triple matching}, integrating both structural and textual features effectively.
A verification step improves alignment results by detecting and correcting misaligned entity pairs.
This step evaluates alignment confidence and consistency, correcting errors by examining the semantic relevance of linearized neighbor triples.
This \textit{Align-then-Verify} approach effectively handles misalignments caused by non-isomorphic structures and noise using rich textual context.

Extensive experiments on five datasets show \proposed outperforms existing methods, demonstrating superior robustness to noisy textual features.
Our verification step achieves near-perfect alignment even with noisy features.
This is significant for knowledge-oriented tasks like EA, where even small accuracy improvements are meaningful.
Furthermore, \textit{Align-then-Verify} pipeline is broadly applicable to other EA methods, enhancing their robustness. 
For reproducibility, our code and datasets are publicly available at \href{https://github.com/SoojinY/ERAlign}{https://github.com/SoojinY/ERAlign}.

The contributions of this paper are summarized below.
\begin{itemize}
    \item \proposed jointly aligns entities and relations using dual graphs and refines them iteratively through \textit{neighbor triple matching}, using both structural and textual information.
 
    \item \proposed introduces the \textit{Align-then-Verify} pipeline, enhancing accuracy even under noisy text features by examining the semantic relevance of neighbor triples. 
    Its robustness to noise is validated through experiments.

    \item \proposed is broadly applicable to other EA methods, enhancing their overall performance and robustness. 
    This versatility makes \proposed a valuable solution, significantly contributing to accuracy and effectiveness and ensuring near-perfect alignment in real-world scenarios.
    
\end{itemize}

%% file: 020related-work.tex
\subsection{Unsupervised Entity Alignment}
\label{subsec:kgalign}
Most EA studies rely on supervised framework using translation embeddings~\cite{transe, mtranse, jape, bootea} or GNN encoders~\cite{gcn-align, rdgcn, hgcn}, 
which heavily depend on manually curated entity pairs that are both costly and error-prone to obtain.
Recently, unsupervised EA methods address the lack of pre-aligned pairs by leveraging side information within KGs to identify seed alignments without human-labeled pairs.
For instance, EVA~\cite{eva} employs ``visual pivoting'' to combine visual and structural information, 
MeaFormer~\cite{meaformer} predicts inter-modal relative weights using images,
DualMatch~\cite{dualmatch} applies weighted graph matching for temporal KGs,
and SEU~\cite{seu} compares entities using structure-enhanced textual features.
SEU reformulates the alignment problem as an OT problem, achieving state-of-the-art accuracy and efficiency. 
By adopting SEU as the base method for initial alignment scores, \proposed avoids the complexity of neural networks while advancing cross-lingual EA.

\subsection{Cross-lingual Entity Alignment}
\label{subsec:textalign}
Cross-lingual EA aligns entities across the multilingual KGs,
where the embedding quality is crucial.
Recent studies enhance embeddings with textual information, such as entity names~\cite{rdgcn, gm-align, hgcn, DualAMN, EASY}, entity descriptions~\cite{KDCoE}, or both~\cite{BERT-INT, RAEA}, 
but often struggle with OOV terms not present in the training data. 
To address this, some methods integrate lexical features like character occurrences~\cite{AttrGNN, EPEA, CEA}, and combine semantic and lexical features~\cite{seu}.
However, most existing approaches focus on textual entity information, overlook \textit{heterogeneous} relationships in KGs.
Relation-focused methods~\cite{srea, raga} address this, but they neglect entity-specific information and rely solely on structural features,
which degrades performance on structurally diverse KGs.
Our goal is to enhance these methods by strategically incorporating textual information from various types of relations,
enabling more accurate alignment by considering both relational structure and textual features.

%% file: 020preliminary.tex
\subsection{Notations and Problem Formulation}
\smallsection{Notations}
KG is represented as $\mathcal{G = (E, R, T)}$, 
where $\mathcal{E}$, $\mathcal{R}$, and $\mathcal{T}$ are set of entities, relations, and triples. 
Each triple is donote as $(h, r, t)\in\mathcal{T}$, 
where $h, t \in \mathcal{E}$ are the head and tail entities, and $r\in \mathcal{R}$ is the relation between them.  
The structural information is captured by the adjacency matrix $\bm{A} \in \mathbb{R}^{|\mathcal{E}|\times |\mathcal{E}|}$.
The textual features of all entities and relations are encoded in $\bm{H}^{ent}\in \mathbb{R}^{|\mathcal{E}|\times d}$ and $\bm{H}^{rel}\in \mathbb{R}^{|\mathcal{R}|\times d}$ where $d$ is the dimensionality.

\smallsection{Problem Formulation}
Cross-lingual EA aims to identify every pair of equivalent entities between the source KG, 
$\mathcal{G}_S=(\mathcal{E}_S, \mathcal{R}_S, \mathcal{T}_S)$, 
and target KG, 
$\mathcal{G}_T=(\mathcal{E}_T, \mathcal{R}_T, \mathcal{T}_T)$, 
which are constructed in different languages. 
In this work, we maintain an \textit{unsupervised} setting, 
where no labeled data is available.

\smallsection{Adjacency Matrix} 
We construct the adjacency matrix to use structural information, following \cite{seu}.
Edges with lower frequency in triples are assigned higher weights, as they convey more unique information. 
The $(i, j)$-th entry of adjacency matrix $\bm{A}$, representing the connection between node $i$ and node $j$, is formulated as:
\begin{equation}
\label{eq:adjmat}
    [\bm{A}]_{ij} = \frac{
    \sum_{r \in\mathcal{R}{(i,j)}} \log{(|\mathcal{T}|/|\mathcal{T}_{r}|)}}
    {\sum_{k \in\mathcal{N}(i)} \sum_{r' \in {\mathcal{R}{(i,k)}}} \log{(|\mathcal{T}|/|\mathcal{T}_{r'}|)}},
\end{equation}
where $\mathcal{N}{(i)}$ is the set of neighboring nodes of node $i$, 
$\mathcal{R}{(i,j)}$ is the set of relations connecting nodes $i$ and $j$,
$\mathcal{T}=\mathcal{T}_S\cup\mathcal{T}_T$ is the set of all triples from the source and target KGs,
and $\mathcal{T}_{r}$ is the set of triples containing relation $r$.
The adjacency matrices are constructed separately for the source and target graphs.

\smallsection{Textual Features} 
The basic distinguish features in a cross-lingual EA dataset are the names of entities and relations.
While some studies enhance representations with entity descriptions \cite{KDCoE, BERT-INT, RAEA}, 
such information is often sparse or unavailable.
To ensure generality, we used only the names as textual features.

Two types of text feature encoding were used for entity and relation: \textit{semantic} and \textit{lexical} features.
In general, {semantic} features are obtained in two ways:
(1) encoding machine-translated text with pre-trained language models (PLMs) 
or (2) capturing semantics through multi-lingual PLMs directly.
However, {semantic} features often face an OOV issue,
especially with proper nouns like celebrity or city names in encyclopedic KGs (e.g., \dbp). 
To address this, we incorporate character-level {lexical} features from translated texts as a complement to semantic features.

\subsection{Optimal Transport-based Entity Alignment}
The innovative approach to unsupervised cross-lingual EA formulates the alignment task as an OT problem~\cite{seu}.
The goal is to match two probability distributions by assigning one to the other to maximize profits, as follows:
\begin{equation}
\label{eq:ot}
    \underset{\bm{P}}{\text{argmax}} \ \langle \bm{P,X} \rangle_F,
\end{equation}
where $\bm{X}$ is the profit matrix, $\bm{P}$ is the transportation matrix, and $\langle \cdot \rangle_F$ is the Frobenius inner product.
In cross-lingual EA task, the transportation matrix $\bm{P}$ must have only one entry of one per each row and column, known as \textit{assignment} problem.

\textit{Sinkhorn}~\cite{sinkhorn} algorithm solves OT problems efficiently, with a time complexity of $O(kn^2)$, as follows:
\begin{align}\label{}
\begin{split}
    S_0(\bm{X}) &= \exp(\bm{X}), \\
    S_k(\bm{X}) &= \mathcal{F}_c(\mathcal{F}_r(S_{k-1}(\bm{X}))), \\
    \text{Sinkhorn}(\bm{X}) &= \lim_{k\rightarrow \infty} S_k(\bm{X}),
\end{split}
\end{align}
where $\mathcal{F}_{r/c}(\cdot)$ is normalization operators for rows and columns that divide each element by the sum of its row and column, respectively.
Sinkhorn algorithm is proven to solve the assignment problem in \cite{sinkhorn2}.
The previous work~\cite{seu} found that a very small $k$ value is enough to perform the EA task, making its time complexity $O(n^2)$, apparently more time efficient.
Thus, we adopt Sinkhorn algorithm to maintain the unsupervised concept of our cross-lingual EA task.

%% file: 030proposed.tex
In this section, we present \proposed, a novel unsupervised cross-lingual EA pipeline.
\proposed is designed to be robust against noisy features, aligning the source and target KGs by leveraging textual features of entities and relations (i.e., names) alongside structural information (i.e., neighbor triples).

\subsection{Overview}
\label{subsec:overview}

\begin{algorithm}[t]
    \caption{The overall process of \proposedpp}
    \label{alg:eralign}
    \SetAlgoLined
    \textbf{Input}: The source and target KGs $\mathcal{G}_S=(\mathcal{E}_S, \mathcal{R}_S, \mathcal{T}_S)$ and $\mathcal{G}_T=(\mathcal{E}_T, \mathcal{R}_T, \mathcal{T}_T)$, the PLM fine-tuned for STS $f_\phi$\\
    \vspace{5pt}
    \tcp{\small Step 1: Dual alignment of entities and relations}
    \nl Construct the dual graphs $\widebar{\mathcal{G}}_S$ and $\widebar{\mathcal{G}}_T$ from ${\mathcal{G}}_S$ and ${\mathcal{G}}_T$ \\
    \nl Obtain the entity alignment matrix $\widehat{\bm{X}}^{ent}$ by Eq.~\eqref{eq:seuoriginal}, \eqref{eq:seuentfinal} \\
    \nl Obtain the relation alignment matrix $\widehat{\bm{X}}^{rel}$ by Eq.~\eqref{eq:seudual}, \eqref{eq:seurelfinal}\\\vspace{5pt}
    \tcp{\small Step 2: Iterative refinement of dual alignment}
    \For{$n\in\{1,\ldots,N\}$}
    {
    \nl Update the alignment matrices ${\bm{S}}_{n}^{ent}$, ${\bm{S}}_{n}^{rel}$ by Eq.~\eqref{eq:fusionent}, \eqref{eq:fusionrel}
    }
    \nl Obtain the alignment matrices $\widehat{\bm{S}}^{ent}$, $\widehat{\bm{S}}^{rel}$ by Eq.~\eqref{eq:entfinal}, \eqref{eq:relfinal}\\
    \vspace{5pt}
    \tcp{\small Step 3: Erroneous alignment verification}
    \For{$i\in\mathcal{E}_S$}{
    \nl Calculate $\textbf{Conf}(i)=\max([\widehat{\bm{S}}^{ent}]{i:})$ \\
    \nl Calculate $\textbf{Cons}(i) = \text{cos}([\widehat{\bm{X}}^{ent}]_{i:}, [\widehat{\bm{S}}^{ent}]_{i:})$
    }
    \nl Collect the entity set for verification $\mathcal{E}^{ver}=\{i|\textbf{Conf}(i)<\theta_{conf} \lor \textbf{Cons}(i)<\theta_{cons}, \forall i\in\mathcal{E}_S\}$ \\
    \For{$i\in\mathcal{E}^{ver}$}{
    \nl Obtain the candidate entities $\text{Cand}(i)=\text{Top-}K([\widehat{\bm{S}}^{ent}]_{i:})$ \\
    \nl Rerank the entities in Cand$(i)$ with $f_\phi$ by Eq.~\eqref{eq:verfsrc}, \eqref{eq:verftgt}
    }
\end{algorithm}

The key idea of \proposed is to simultaneously align entities and relations across the source and target KGs, through iterative fusion of entity-level and relation-level alignment using information within the KGs.
\proposed consists of the three steps: 
(1) dual alignment of entities and relations, 
(2) iterative refinement of dual alignment, and 
(3) erroneous alignment verification.

First, \proposed computes the initial entity-level and relation-level alignment scores using textual features extracted from the original KGs (for entities) and dual KGs (for relations).
It then iteratively refines these alignments via \textit{neighbor triple matching}, aligning the relevant triples between the source and target KGs. 
Finally, it detects and corrects misaligned pairs using language models to directly capture the semantic meaning of the neighbor triples.
Algorithm~\ref{alg:eralign} outlines the pseudo-code for \proposed. 



\subsection{Dual Alignment of Entities and Relations}
\label{subsec:stepone}
\proposed computes alignment scores between $\mathcal{G}_S$ and $\mathcal{G}_T$ using entity-level and relation-level similarity matrices using textual features.
Entity features are enhanced by propagating textual features to their $l$-hop neighbor using a graph convolution operation, $(\bm{A}^{ent})^l\bm{H}^{ent}$ for depth of $l$.
The entity-level similarities are determined by computing the inner product of these structure-enhanced entity features.
The final similarity matrix $\bm{X}^{{ent}}\in\mathbb{R}^{|\mathcal{E}_S|\times|\mathcal{E}_T|}$ is the sum of the similarities across depths $l=0,\ldots,L$:
\begin{align}
\label{eq:seuoriginal}
    \bm{X}^{{ent}} &= \sum^{L}_{l=0} \left((\bm{A}^{{ent}}_{S})^l\bm{H}_{S}^{{ent}}\right)\left((\bm{A}_{T}^{{ent}})^l\bm{H}_{T}^{{ent}}\right)^{\top}.
\end{align}


We also construct a \textit{dual knowledge graph} to enhance relation features through textual propagation,
$\widebar{\mathcal{G}}=(\widebar{\mathcal{E}}, \widebar{\mathcal{R}}, \widebar{\mathcal{T}})$, 
derived from the original KG $\mathcal{G}$.
In this dual KG, nodes represent edges in $\mathcal{G}$ (i.e., $\widebar{\mathcal{E}}={\mathcal{R}}$) and edges correspond to nodes in $\mathcal{G}$ (i.e., $\widebar{\mathcal{R}}={\mathcal{E}}$).\footnote{
As the edges in the dual KGs lack directionality, the dual KGs are represented as undirected graphs by including two symmetric triples $(r, e, r')$ and $(r', e, r)$.
}
The adjacency matrix of the dual KG, $\bm{A}^{rel}$, is constructed similarly to Equation~\eqref{eq:adjmat}.
Utilizing dual KGs, the relation-level similarity matrix $\bm{X}^{{rel}}\in\mathbb{R}^{|\mathcal{R}_S|\times|\mathcal{R}_T|}$ is defined similarly to Equation~\eqref{eq:seuoriginal}, ensuring consistency in propagating textual features through the dual KGs:
\begin{align}
\label{eq:seudual}
    \bm{X}^{{rel}} &= \sum^{L}_{l=0} \left((\bm{A}^{{rel}}_{S})^l\bm{H}_{S}^{{rel}}\right)\left((\bm{A}^{{rel}}_{T})^l\bm{H}^{{rel}}_{T}\right)^{\top},
\end{align}

Applying the Sinkhorn algorithm to the similarity matrices produces the initial alignment results.
This process effectively establishes a correspondence between nodes across both the original and dual KGs based on their computed similarity.
\begin{align}
\label{eq:seuentfinal}
    \widehat{\bm{X}}^{ent} &= \lim_{\tau \rightarrow 0^+} \text{Sinkhorn}(\bm{X}^{ent} /\tau),\\
\label{eq:seurelfinal}
    \widehat{\bm{X}}^{rel} &= \lim_{\tau \rightarrow 0^+} \text{Sinkhorn}(\bm{X}^{rel} /\tau).
\end{align}

Considering effectiveness and unsupervised setting, we used \cite{seu} approach as our base method. 
Other methods could also be chosen as the base method to obtain initial alignment scores,
since our dual KG approach allows for obtaining relation-level alignment scores in the same manner as obtaining entity-level alignment scores.

\begin{figure}[t]
    \centering
    \includegraphics[width=\linewidth]{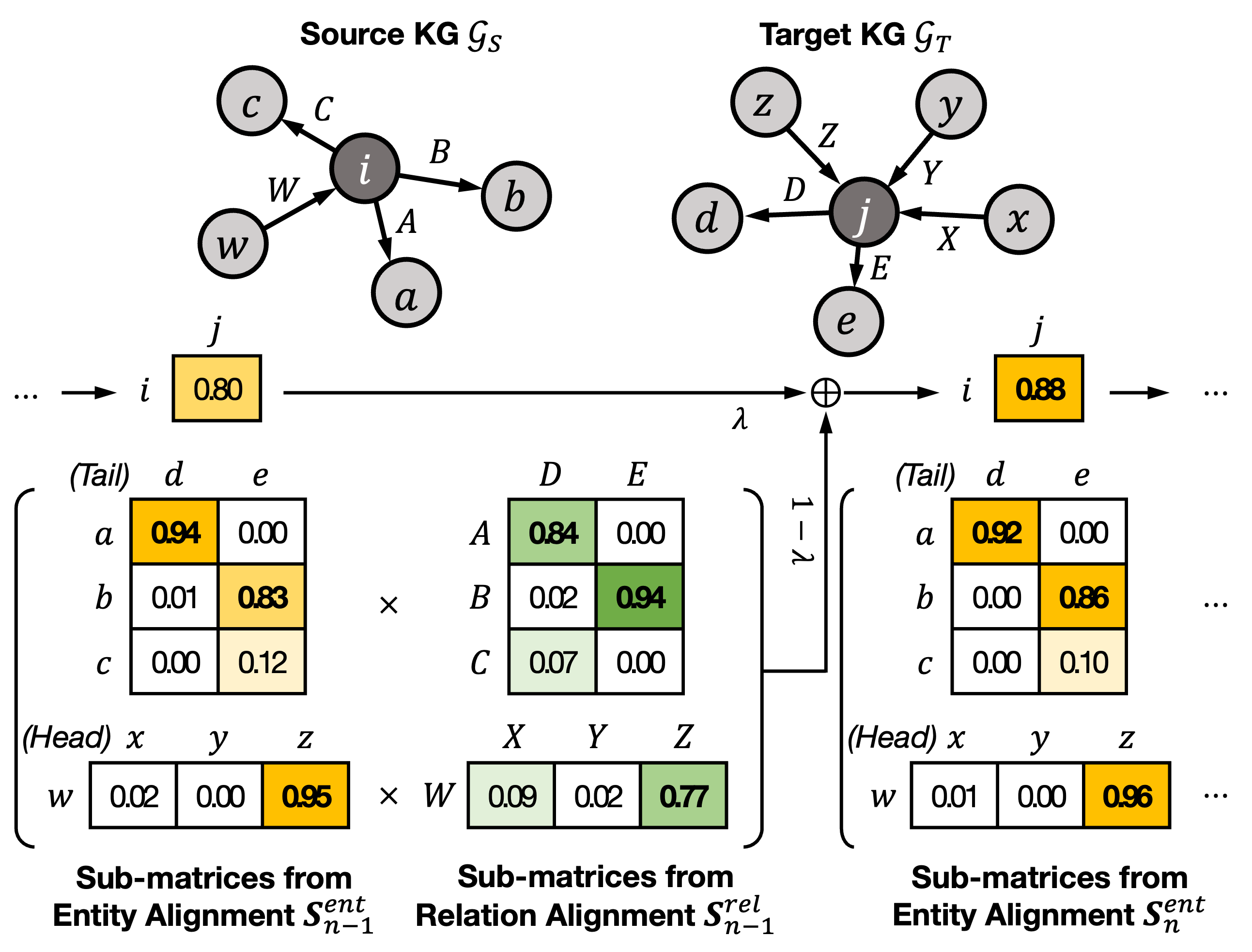}
    \caption{Iterative refinement of $(i,j)$ entity alignment based on neighbor triple matching. Each relation alignment is refined in the same way by using the dual KGs.}
    \label{fig:refine}
\end{figure}

\subsection{Iterative Refinement of Dual Alignment}
\label{subsec:steptwo}
In the second step, \proposed iteratively refines the alignment scores for both entity-level and relation-level alignment by integrating their align results through \textit{neighbor triple matching} between the source and target KGs (Figure~\ref{fig:refine}).
The initial alignment scores for entities, $\bm{S}^{ent}_{0}$, and relations, $\bm{S}^{rel}_{0}$, are derived from the matrices $\widehat{\bm{X}}^{ent}$ and $\widehat{\bm{X}}^{rel}$, as specified in Equations~\eqref{eq:seuentfinal} and \eqref{eq:seurelfinal}, respectively.

For the $(i, j)$-th entity alignment, scores are refined by aggregating alignment scores from neighboring triples in $\mathcal{G}_S$ and $\mathcal{G}_T$. 
Specifically, all neighbor triples  
$(i, p, i') \in\mathcal{T}_S$ and $(j, q, j')\in\mathcal{T}_T$ (or $(i', p, i) \in\mathcal{T}_S$ and $(j', q, j)\in\mathcal{T}_T$) are used,
weighting tail (or head) entity pairs $(i', j')$ by the corresponding relation pair $(p, q)$.
Neighbor triples of entity $i$  are identified by checking whether $i$ is the head (or tail) entity in each triple $(h, r, t)$.
Similarly, $(i, j)$-th relation alignment score is updated using neighbor triples in the dual KGs. 
The $n$-th iteration updates are defined as:
\begin{align}
\label{eq:fusionent}
[\bm{S}^{ent}_{n}]_{ij} = \lambda &\cdot [\bm{S}^{ent}_{n-1}]_{ij} \\ 
\nonumber
+ (1-\lambda) &\cdot \hspace{-5pt} \sum_{\substack{(i,p,i')\in\mathcal{T}_{S}'\\(j,q,j')\in\mathcal{T}_{T}'}}
\hspace{-5pt} [\bm{S}^{ent}_{n-1}]_{i'j'} \cdot \frac{\exp ([\bm{S}^{rel}_{n-1}]_{pq})}{ \sum_{\substack{(i,p,i')\in{\mathcal{T}}_{S}'\\(j,q,j')\in{\mathcal{T}}_{T}'}} \exp ([\bm{S}^{rel}_{n-1}]_{pq})}, \\ 
\label{eq:fusionrel}
[\bm{S}^{rel}_{n}]_{ij} = \lambda &\cdot [\bm{S}^{rel}_{n-1}]_{ij} \\ 
\nonumber
+ (1-\lambda) &\cdot \hspace{-5pt} \sum_{\substack{(i,p,i')\in\widebar{\mathcal{T}}_{S}\\(j,q,j')\in\widebar{\mathcal{T}}_{T}}}
\hspace{-5pt} [\bm{S}^{rel}_{n-1}]_{i'j'} \cdot \frac{\exp ([\bm{S}^{ent}_{n-1}]_{pq})}{  \sum_{\substack{(i,p,i')\in\widebar{\mathcal{T}}_{S}\\(j,q,j')\in\widebar{\mathcal{T}}_{T}}} \exp ([\bm{S}^{ent}_{n-1}]_{pq})}, 
\end{align}
where $\lambda\in(0,1)$ is the hyperparameter to govern the weighting of the alignment score from the preceding iteration.
For notational convenience, we define the reverse-augmented triple set as $\mathcal{T}':=\mathcal{T}\cup\{(t, r, h)|(h, r, t)\in\mathcal{T}\}$, representing all neighboring triples of an entity.
At each iteration, entity and relation alignments are intertwined through 1-hop neighbor triple matching.
By increasing the total number of iterations $N$, \proposed incorporates a more extensive relational structure within the KGs.

The final alignment results for the EA task are obtained by applying the Sinkhorn algorithm to the alignment scores $\bm{S}^{ent}_{N}$ and $\bm{S}^{rel}_{N}$.
Sinkhorn algorithm leads to more accurate and reliable cross-lingual entity and relation alignments. 
\begin{align}
\label{eq:entfinal}
    \widehat{\bm{S}}^{ent} &= \lim_{\tau \rightarrow 0^+} \text{Sinkhorn}(\bm{S}^{ent}_{N} /\tau),\\
\label{eq:relfinal}
    \widehat{\bm{S}}^{rel} &= \lim_{\tau \rightarrow 0^+} \text{Sinkhorn}(\bm{S}^{rel}_{N} /\tau).
\end{align}

\subsection{Erroneous Alignment Verification}
\label{subsec:stepthree}
The final step involves verification to enhance robustness against noisy textual features, which often arise from machine translation or encoding issues 
and significantly impact cross-lingual EA performance.
To mitigate this challenge, we analyze the neighbor triples of each entity as linearized text,
a strategy less affected by noise within rich textual contexts. 
Our verification step identifies flawed entity alignments then reorders the entities according to their semantic similarity computed by PLMs.
\footnote{Following the standard evaluation protocol for cross-lingual EA, our verification process focuses on reordering aligned entities within the target KG for each source KG entity.}

\subsubsection{Detection of erroneous alignments}
\label{subsubsec:detection}
To effectively identify misaligned entity pairs, we devise and examine two distinct metrics tailored to assess the accuracy of entity alignment. 
These metrics are individually crafted for each entity within the source KG, $i\in\mathcal{E}_S$.
\begin{itemize}
    \item \textbf{Confidence}: 
    It gauges the confidence with which the top-1 entity (in $\mathcal{E}_T$) is aligned with $i$:
    $\textbf{Conf}(i)=\max([\widehat{\bm{S}}^{ent}]_{i:})$.
    
    \item \textbf{Consistency}: 
    It reflects how consistently entities (in $\mathcal{E}_T$) are aligned during iterative refinement.
    Higher consistency implies less discrepancy between the entity-level and relation-level alignments:
    $\textbf{Cons}(i) = \text{cos}([\widehat{\bm{X}}^{ent}]_{i:}, [\widehat{\bm{S}}^{ent}]_{i:})$.
\end{itemize}

\begin{figure}[t]
    \centering
        \includegraphics[width=0.49\linewidth]{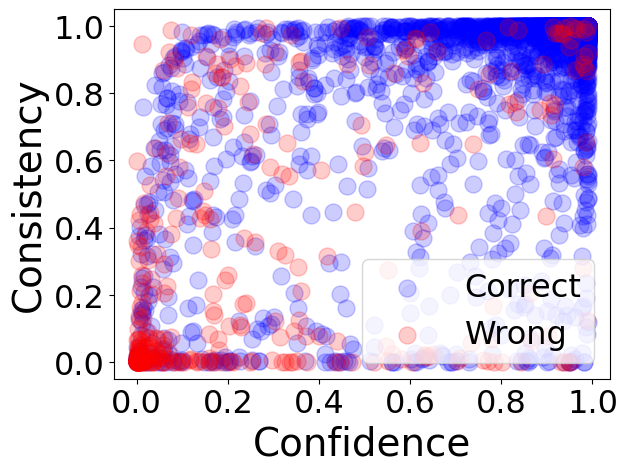}
        \includegraphics[width=0.49\linewidth]{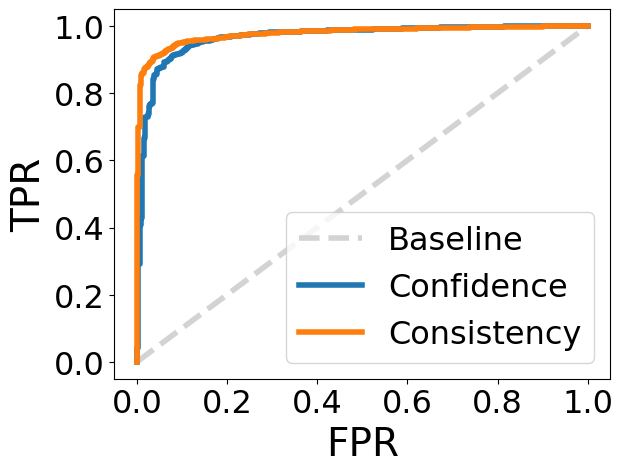}
    \caption{Confidence and consistency scores for each entity (Left). AUROC of the two metrics in detecting errors (Right).}
    \label{fig:detection}
\end{figure}

We conduct a thorough analysis of the correlation between these metrics and the actual alignment errors made by \proposed. 
In Figure~\ref{fig:detection}, it is evident that erroneous entity pairs are distinguishable from correct pairs based on their confidence and consistency. 
Precisely, the AUROC (and AUPR) of confidence and consistency for error detection are respectively 0.965 (0.998) and 0.977 (0.999), indicating the effectiveness of these metrics as discriminators.
In light of this, we selectively collect the set of entities for verification, denoted as $\mathcal{E}^{ver}=\{i|\textbf{Conf}(i)<\theta_{conf} \lor \textbf{Cons}(i)<\theta_{cons}, \forall i\in\mathcal{E}_S\}$. 

\subsubsection{Correction of erroneous alignments}
\label{subsubsec:correction}
After collecting the potentially incorrect entities, we replace their alignment scores with semantic relevance derived from the linearized texts of the neighbor triples associated with these entities.
For each entity $i\in\mathcal{E}^{ver}$, we identify its top-$K$ aligned entities as candidates, denoted as $\text{Cand}(i) = \text{Top-}K([\widehat{\bm{S}}^{ent}]_{i:})$. 
It is these candidates that undergo reranking based on textual semantic relevance.
\footnote{For computational efficiency, we fix the number of candidate entities to $K=20$. The detailed analysis is provided in Section~\ref{subsubsec:paramanal}.}
To measure textual semantic relevance, we utilize PLMs~\cite{bert,roberta,minilm} fine-tuned specifically for the task of semantic textual similarity (STS).

The semantic relevance scores obtained from PLMs hinge on the order of neighbor triples within the linearized text.
We initially sort all neighbor triples $(i,p,i')\in\mathcal{T}_{S}'$ for an entity $i\in\mathcal{E}^{ver}$ based on the inverse frequency of its relation $p$ (Equation~\eqref{eq:adjmat}). 
Subsequently, we sort the neighbor triples $(j,q,j')\in\mathcal{T}_{T}'$ of each candidate entity $j\in\text{Cand}(i)$ so that their order becomes consistent with the order of $(i,p,i')$ based on the alignment score, i.e., $[\widehat{\bm{S}}^{rel}]_{pq}\cdot[\widehat{\bm{S}}^{ent}]_{i'j'}$.
The resulting sequence of neighbor triples is then directly transformed into linearized text, serving as the basis for reranking entities.
\begin{align}
    \label{eq:verfsrc}
    \mathcal{V}_{T}(i) &=\text{argsort}_{j\in\text{Cand}(i)}(f_\phi(Z_S(i), Z_T(j))), \\
    \label{eq:verftgt}
    \mathcal{V}_{S}(j) &=\text{argsort}_{i\in\text{Cand}(j)}(f_\phi(Z_S(i), Z_T(j))), 
\end{align}
where $Z_S(i)$ and $Z_T(j)$ are the linearized texts of entities $i$ and $j$'s neighbor triples.
The function $f(\cdot,\cdot)$ calculates the relevance, and $\phi$ denotes the parameters of the PLM.
In the end, we can obtain the reordered list $\mathcal{V}$ of candidate entities based on semantic textual similarity.
For linearization, we use a straightforward text template that summarizes relevant knowledge, focusing on the entity.
For instance, the entity ``Football League One'' can be represented through linearized text based on its neighbor triples:
``Football League One, which relegation is Football League Championship, promotion is Football League Championship, promotion is Football League Two, relegation is Football League Two, league is Sheffield United F.C..''.


In addressing the correction of the entity pair $(i, j)$, our empirical findings indicate that the reranking strategy, as applied to entities aligned with $i\in\mathcal{G}_S$ (Equation~\eqref{eq:verfsrc}), falls short of fully harnessing the ranking knowledge associated with entities aligned to $j\in\mathcal{G}_T$.
To overcome this limitation, we adopt a \textit{cross-verification} strategy.
This approach entails acquiring the ranked list for entity $j\in\mathcal{G}_T$ (Equation~\eqref{eq:verftgt}) and selectively replacing alignment outcomes only when the conditions $j==\mathcal{V}_T(i)_1$ and $i==\mathcal{V}_S(j)_1$ are satisfied.

%% file: 040experiments.tex

\subsection{Experimental Settings}
\begin{table}[t]
    \caption{Statistics of datasets for cross-lingual EA.}
    \label{tbl:datastats}
    \centering
    \resizebox{0.99\linewidth}{!}{
        \begin{tabular}{XWccc}
            \toprule
            \multicolumn{2}{c}{\textbf{Dataset}} & \textbf{\#Entities} & \textbf{\#Relations} & \textbf{\#Triples}\\
            \midrule
            
            \multirow{3}{*}{\rotatebox[origin=c]{0}{\footnotesize \dbp}}
            & {\scriptsize {ZH-EN}} 
            & {19,388/19,572} & {1,707/1,323} & {\ \ 70,414/\ \ 95,142}\\
            & {\scriptsize {JA-EN}} 
            & {19,814/19,780} & {1,299/1,153} & {\ \ 77,214/\ \ 93,484}\\
            & {\scriptsize {FR-EN}} 
            & {19,661/19,993} & {\ \ \ 903/1,208} & {105,998/115,722}\\
            \midrule
            
            \multirow{2}{*}{\rotatebox[origin=c]{0}{\small \srp}} 
            & {\scriptsize {EN-FR}} 
            & {15,000/15,000} & {\ \ \ 177/\ \ \ 221} &{ \ \ 33,532/\ \ 36,508}\\
            & {\scriptsize {EN-DE}} 
            & {15,000/15,000} & {\ \ \ 120/\ \ \ 222} & {\ \ 37,377/\ \ 38,363}\\
            \bottomrule
        \end{tabular}
        }
\end{table}

\smallsection{Datatsets}
We use five public datasets, extensively used in previous works~\cite{seu, DualAMN, hgcn}: 
\textbf{\dbp}~\cite{dataset_dbp15k} are the three cross-lingual subsets from multi-lingual DBpedia.
\textbf{\srp}~\cite{dataset_srprs} cover much sparse KGs.
The statistics of the datasets are provided in Table~\ref{tbl:datastats}.

\smallsection{Baseline methods}
Our main goal is to demonstrate how effectively \proposed enhances the base method using semantic information via \textit{neighbor triple matching} as a general pipeline, rather than focusing solely on achieving state-of-the-art performance.
Thus, to ensure a fair comparison, the selection of baselines prioritized the level of entity information utilized~\cite{meaformer} over the use of the latest methods.
While \proposed mainly utilizes entity names for semantic information, other recent studies~\cite{Jiang2023UnsupervisedDC, peea, fgwea} integrate attributes and other features to enhance entity representations.

We categorize methods by feature encoding strategies.
\textit{Structure-based methods} use only structure information: {MTransE}~\cite{mtranse}, {GCN-Align}~\cite{gcn-align}, and {BootEA}~\cite{bootea}.
\textit{Semantic-based methods} combine textual and structure information: {JEANS}~\cite{jeans}, {GM-Align}~\cite{gm-align}, {HGCN}~\cite{hgcn}, and {Dual-AMN}~\cite{DualAMN}.
\textbf{\proposed} is our proposed pipeline that jointly refines the entity and relation alignment through neighbor triple matching in an iterative manner.
\textbf{\proposedpp} extends this with an additional verification step to detect and correct alignment errors, forming an \textit{Align-then-Verify} pipeline.
To demonstrate \proposed as a versatile pipeline for enhancing base methods, we adopt three unsupervised (or self-supervised) methods for initial alignment scores:
{\seu}~\cite{seu}, {\lightea}~\cite{lightea}, and {\eva}~\cite{eva}.

Most baseline methods are supervised and rely on pre-aligned seed pairs,
typically using 30\% for training (and development) and 70\% for testing.
In contrast, \seu and \lightea are parameter-free unsupervised frameworks that test on all entity pairs.
For consistency, we exclude multi-modal EA methods that incorporate visual or temporal information within KGs~\cite{meaformer, dualmatch}.

\smallsection{Implementation details}
For \proposed, alignment scores from SEU are used as the default initial scores (Section~\ref{subsec:stepone}).
To encode text information, we employ Sentence BERT (SBert)\footnote{LaBSE, available at https://huggingface.co/sentence-transformers/LaBSE.} for \textit{semantic features} and character-level bigram (Bigram) for \textit{lexical features}.
In the iterative refinement process (Section~\ref{subsec:steptwo}), the hyperparameters are fixed to $\lambda=0.5$ and $N=2$.
For erroneous alignment detection (Section~\ref{subsubsec:detection}), thresholds $\theta_{conf}$ and $\theta_{cons}$ are configured to ensure that 20\% of source entities undergo additional verification.
\footnote{The detection process aims to enhance efficiency and effectiveness by identifying entities requiring verification, rather than achieving precise discrimination. 
Further refinement in accurately detecting erroneous pairs is left for future studies.}
For the correction of erroneous alignments (Section~\ref{subsubsec:correction}), sentence transformer\footnote{{https://huggingface.co/sentence-transformers/all-MiniLM-L6-v2}.} measures textual semantic relevance, considering $K=20$ candidate entities for reranking.

\subsection{Experimental Results}
\subsubsection{Effectiveness of \proposed}
\label{subsubsec:effectrefine}

\input{041maintable}

For comprehensive comparisons, we evaluate \proposed, \proposedpp, and baseline methods.
As shown in Table~\ref{tbl:main}, \proposed and \proposedpp achieve the highest accuracy on most datasets without relying on human-curated pairs or learnable parameters.
Structure-based methods perform worse than those using textual features.
Supervised EA methods often suffer from overfitting due to the small size of training data (i.e., pre-aligned pairs),
making the unsupervised approach, \seu and \proposed, more effective with their Sinkhorn which mitigates overfitting.
\proposed consistently outperforms its base method (i.e., \seu) in accuracy across all datasets, demonstrating the advantage of our iterative refinement based on neighbor triple matching. 
This shows that \proposed is a pipeline that effectively enhances the robustness of the base method.
The improvement is particularly notable for \dbpzh and \dbpja, as these languages belong to distinct language families compared to English;
this contrasts with French and German, which share common linguistic roots with English.
Moreover, \proposedpp elevates Hit@1 scores with its verification step, showing that considering semantic similarity between neighbor triples enhances alignment performance.

\input{042validtable}
\subsubsection{Effect of textual feature types}
\label{subsubsec:effectfeat} 
The accuracy of the dual alignment, $\widehat{\bm{X}}^{ent}$ and $\widehat{\bm{X}}^{rel}$, is assessed in Section~\ref{subsec:stepone} using various textual features (or text encoding strategies), including semantic features (i.e., \textbf{\glove} and \textbf{\sbert}) and lexical features (i.e., \textbf{Bigram}).
To evaluate relation alignment, a subset of aligned relation pairs within the KGs is used.
Table~\ref{tbl:featvalid} shows that alignment performance heavily depends on the quality of textual features for both entities and relations.
\sbert outperforms \glove by better capturing word token contexts, emphasizing the importance of effective semantic encoding.
Moreover, concatenating Bigram features significantly enhances the accuracy of both \glove and \sbert by allowing $\widehat{\bm{X}}^{ent}$ and $\widehat{\bm{X}}^{rel}$ to encode lexical similarities between entities and relations, respectively.
This approach proves highly effective in handling OOV words or phrases within entities and relations.

\subsubsection{Generalizability of \proposed}
\label{subsubsec:application}
To validate that \proposed can serve as a general pipeline for robust cross-lingual EA,
we employ three unsupervised (or self-supervised) EA methods (i.e., \textbf{\eva}, \textbf{\lightea}, and \textbf{\seu}) as the base method for the \proposed pipeline.
The entity-entity (and relation-relation) alignment scores obtained by \eva, \lightea, and \seu are used as the initial alignment scores for \proposed (Step 1).
Note that the proposed refinement and verification steps are entirely independent of the method used to compute these initial scores.  
To ensure consistency, \eva and \lightea are tailored to fit our experimental settings. 
The entire labeled data is used as the test set, with \sbert and Bigram concatenated for textual embeddings, while adhering to the settings in the original papers.

\begin{figure}[t]
    \centering
        \includegraphics[width=1\linewidth]{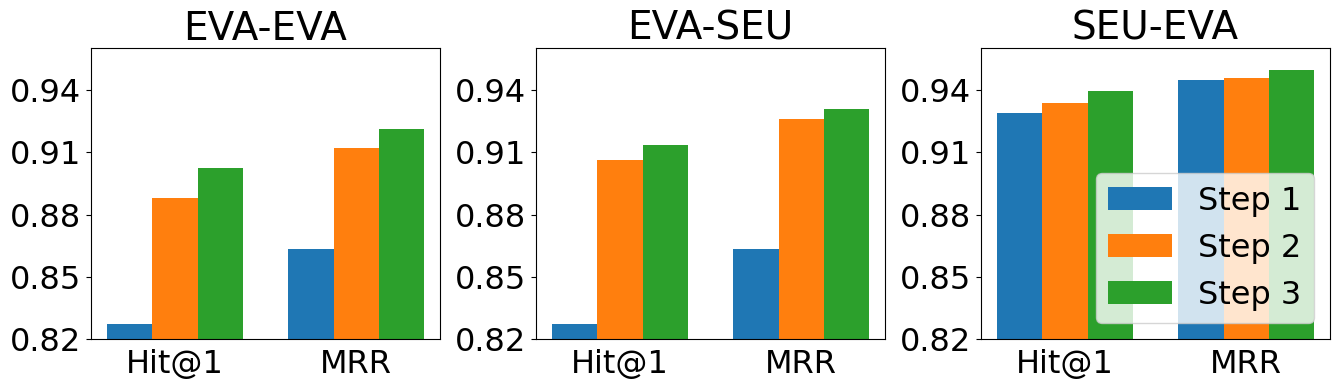}
        \includegraphics[width=1\linewidth]{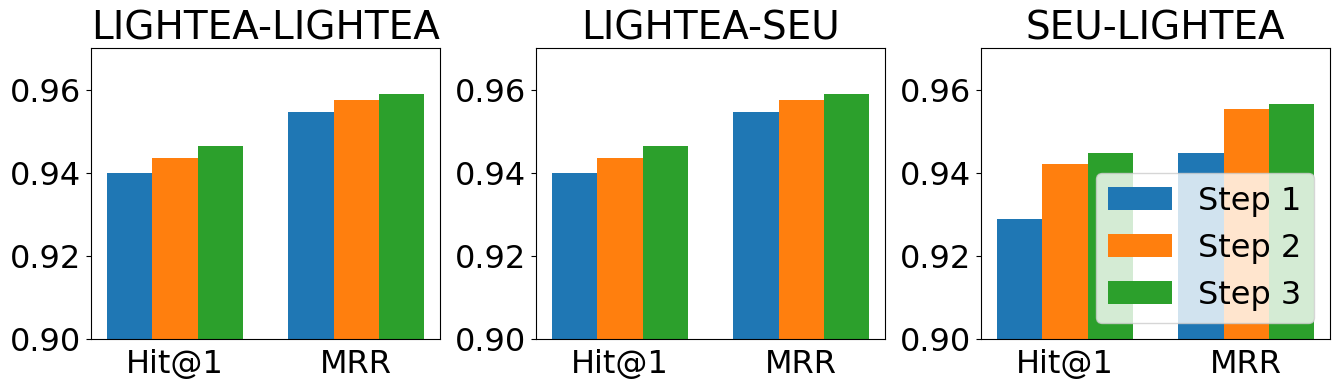}
    \caption{Performance of cross-lingual EA methods in our \proposed pipeline on \dbpzh, with annotated methods for entity-level and relation-level scores.}
    \label{fig:application}
\end{figure}

Figure~\ref{fig:application} illustrates the performance at each step of \proposed;
Step 1 shows the performance of the initial alignment scores,
Step 2 is the performance after the refinement step, 
and Step 3 is the performance of the verification.
The results demonstrate that \proposed is effective, especially when the accuracy of initial entity-entity alignment is low.
That is, the differences in results depend on the initial alignment scores. 
While we exclude other modalities for fairness, incorporating them might further improve the performance of \proposed.
This experiment highlights the broad applicability of \proposed, showing that the base method for initial alignment scores is not restricted to \seu.
It also suggests that future state-of-the-art methods could achieve more robust performance through \proposed.

\input{045dropratio}
\subsubsection{Robustness to heterogeneity between KGs}
\label{subsubsec:drop}
Most EA methods assume that aligned entity pairs share similar structure across source and target graphs.
However, perfectly isomorphic graphs are rare in practice, leading to performance declines due to KG heterogeneity.
To simulate non-isomorphic scenarios, we randomly drop triples from the target graph of \dbpzh at varying rates (0\%, 25\%, 50\%, and 75\%). 
A high drop ratio introduces substantial non-isomorphism between the KGs.
Table~\ref{tbl:dropratio} shows that in highly non-isomorphic scenarios, both Step 1 (Initialization) and Step 2 (Refinement) can introduce noise,
as these steps assume aligned pairs have identical representations, including structure and text.
At a 75\% drop ratio, Step 2 even degrades performance compared to Step 1;
however, Step 3 (Verification) mitigates these detrimental effects.
Despite structural differences, the semantic contexts of entities and relations in neighbor triples remain similar.
Consequently, the semantic similarities of verbalized neighbor triples help improve alignment performance in highly non-isomorphic situations.

\subsubsection{Robustness to noisy textual features}
\label{subsubsec:noise}

\begin{figure}[t]
    \centering
        \includegraphics[width=0.49\linewidth]{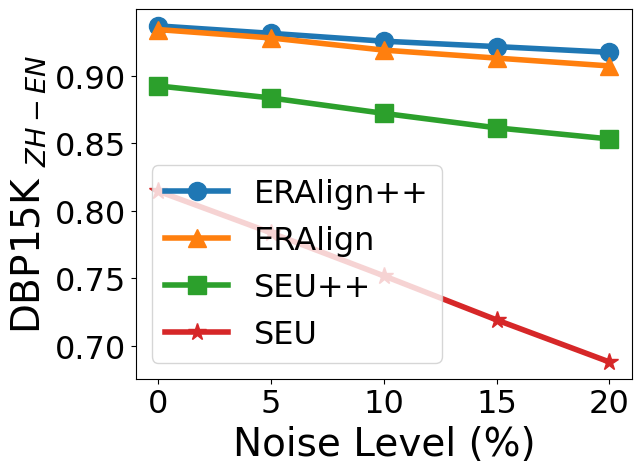}
        \includegraphics[width=0.49\linewidth]{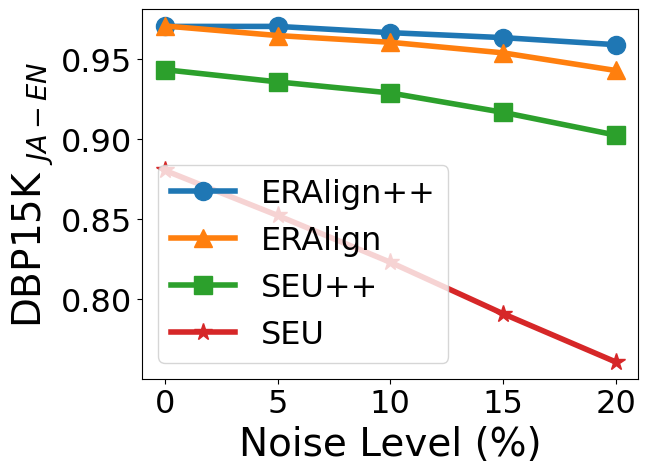}
        \includegraphics[width=0.49\linewidth]{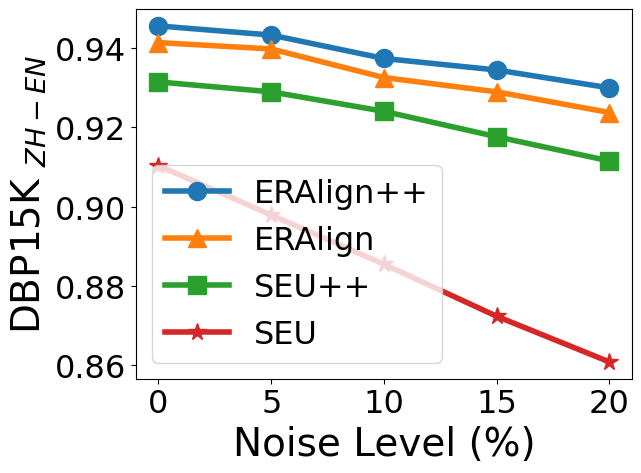}
        \includegraphics[width=0.49\linewidth]{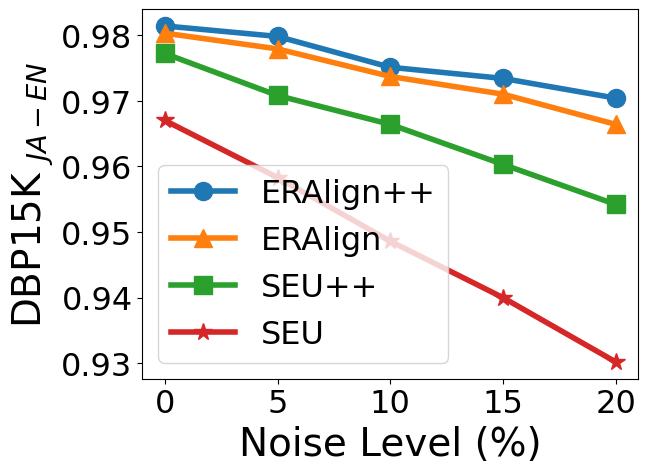}
    \caption{Robustness (Hit@1) of \proposed and \seu to simulated textual noises in entities and relations, with semantic features (Upper) and semantic+lexical features (Lower). \proposedpp and \seupp incorporate our verification step.}
    \label{fig:robustness}
\end{figure}

\input{043casestudy}

The robustness of \proposed and \proposedpp is demonstrated by evaluating their alignment accuracy under artificially injected textual noise levels ranging from 0\% to 20\%. 
The experiments simulate various types of textual noise: phonetic errors, missing characters, and attached characters.
In Figure~\ref{fig:robustness}, \proposed shows lower sensitivity to textual noises compared to \seu, attributed to our iterative refinement step that enhances noise tolerance through the relation-level alignment fusion. 
The incorporation of lexical features consistently improves accuracy by capturing lexical similarities among entities and relations.
Notably, the verification step enhances robustness in both \proposedpp and \seupp,
effectively correcting erroneous alignments through semantic similarity in linearized neighbor triples.
Its effectiveness becomes particularly pronounced as noise increases, strongly indicating that the additional verification substantially enhances alignment accuracy in the presence of textual noise.

\subsubsection{Hyperparameter analysis}
\label{subsubsec:paramanal}

\begin{figure}[t]
    \centering
        \includegraphics[width=1\linewidth]{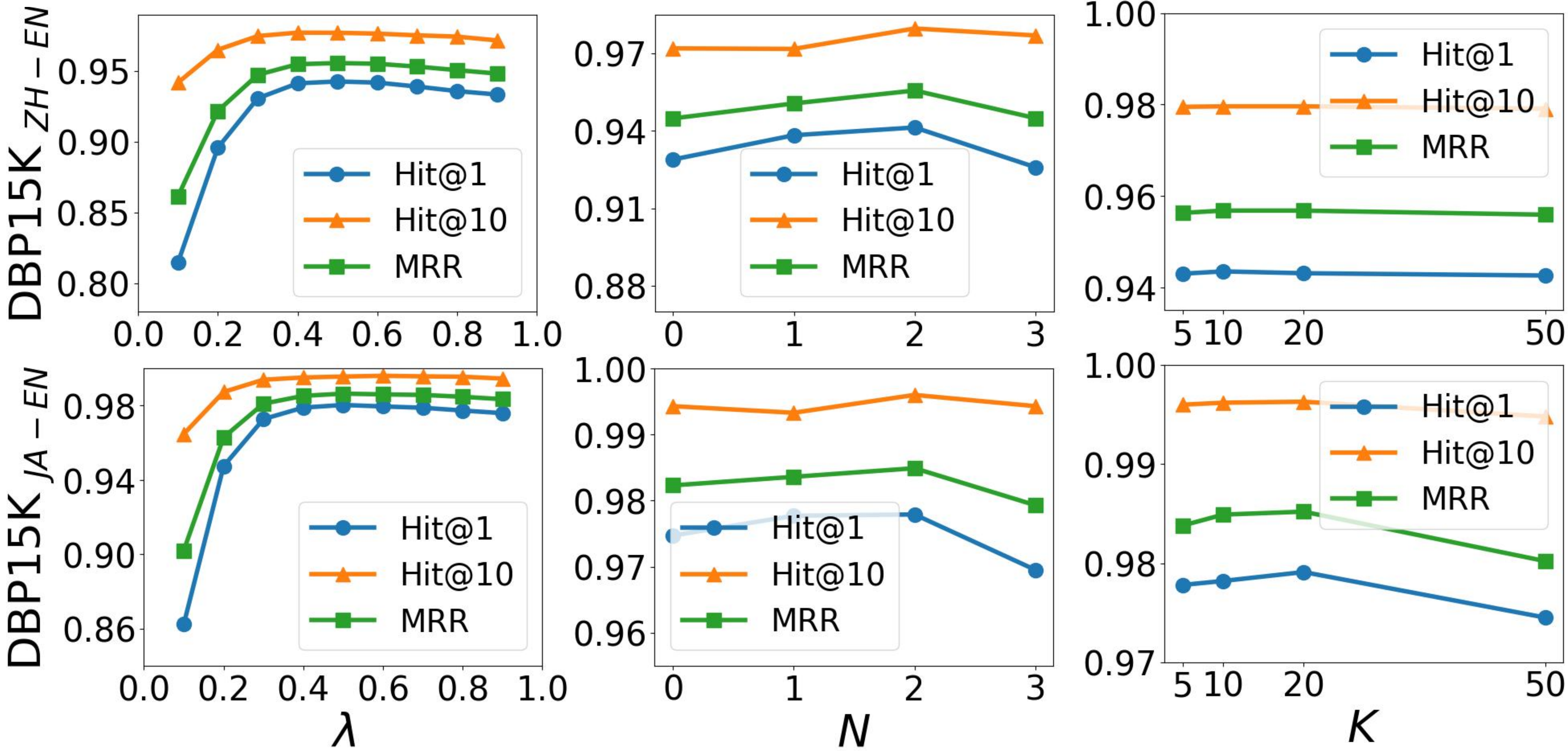}
    \caption{Performance changes of \proposed with respect to the hyperparameters: $\lambda$ (Upper), $N$ (Middle), and $K$ (Lower).}
    \label{fig:paramanal}
\end{figure}

We examine the impact of various hyperparameters on the performance of \proposed using the \dbpzh and \dbpja datasets. 
These hyperparameters include the weight for the previous alignment ($\lambda$), the number of refinement iterations ($N$), and the number of candidate entities ($K$) considered during the verification of erroneously aligned pairs.
As shown in Figure~\ref{fig:paramanal}, \proposed maintains consistent performance for $\lambda$ values in the range [0.3, 0.9]. 
Optimal performance is consistently achieved at $N=2$ for both datasets.
For $K$, values between 10 and 20 are found be optimal.
A smaller $K$ may fail to include the correct entity in the candidate set, while a large $K$ introduces noise, reducing the effectiveness of the correction process.

\subsubsection{Ablation study}
\label{subsubsec:ablation}
To assess the contributions of individual components in \proposed and \proposedpp, ablation analyses are conducted to evaluate their impact on alignment accuracy. 
Variants of \proposed include:
(1) replacing \sbert features with \glove~\cite{glove}, and
(2) using \sbert without Bigram features, compared to \seu, which uses \glove and Bigram. 
For \proposedpp, modifications are made to:
(3) exclude confidence and (4) exclude consistency metrics for erroneous alignment detection,
(5) skip sorting neighbor triples during linearization, and
(6) replace the STS task with an NLI fine-tuned language model for correction.
Table~\ref{tbl:ablation} shows the importance of semantic and lexical features for \proposed. 
For \proposedpp, confidence and consistency metrics are crucial for robust verification, reducing false positives (i.e., correctly aligned pairs). 
Furthermore, sorting triples linearization and using the STS task with the PLM are critical for accurate error correction.

\input{044ablation}

\subsubsection{Case study on erroneous alignment correction}
\label{subsubsec:casestudy}
\begin{CJK*}{UTF8}{gbsn}
An example of the alignment correction process is provided in Table~\ref{tbl:casestudy}, showing how \proposedpp accurately rectifies entity pairs in \dbpzh.
Each entity (and relation) name is annotated with its English translation from the original language.
\proposedpp correctly identifies the target entity ``Gan Chinese'' ($\rightarrow$ ``Gan Chinese'') as most relevant to the source entity ``赣语'' ($\rightarrow$ ``Gan language''). 
This indicates that directly examining neighbor triple sets using semantic textual similarity effectively resolves the error.
\end{CJK*}

%% file: 041maintable.tex
\begin{table*}[t]
\caption{The cross-lingual EA accuracy of \proposed, \proposedpp, and baseline methods. `S', `SS', and `U' denote supervised, semi-supervised, and unsupervised methods, respectively. 
\seu is reproduced by applying Sinkhorn to all entities in KGs, whereas the original paper focused only on entities with aligned pairs, potentially overestimate performance.
}
    \label{tbl:main}
    \centering
    \resizebox{0.99\linewidth}{!}{%
    \begin{tabular}{clcYYYYYYYYYYYYYYY}
        \toprule
        
        & \multirow{2.5}{*}{\textbf{Method}} &
        & \multicolumn{3}{c}{\textbf{\dbpzh}} 
        & \multicolumn{3}{c}{\textbf{\dbpja}} 
        & \multicolumn{3}{c}{\textbf{\dbpfr}} 
        & \multicolumn{3}{c}{\textbf{\srpfr}} 
        & \multicolumn{3}{c}{\textbf{\srpde}} \\ 
        \cmidrule(lr){4-6}\cmidrule(lr){7-9}\cmidrule(lr){10-12}\cmidrule(lr){13-15}\cmidrule(lr){16-18}

        & &
        & \small H@1 & \small H@10 & \small MRR
        & \small H@1 & \small H@10 & \small MRR
        & \small H@1 & \small H@10 & \small MRR
        & \small H@1 & \small H@10 & \small MRR
        & \small H@1 & \small H@10 & \small MRR\\ 
        \midrule
        


        \multirow{3}{*}{\rotatebox[origin=c]{90}{\small Structure}}
        & MTransE & S
        & 0.209 & 0.512 & 0.310 
        & 0.250 & 0.572 & 0.360 
        & 0.247 & 0.577 & 0.360 
        & 0.213 & 0.447 & 0.290
        & 0.107 & 0.248 & 0.160\\ 
        
        & GCN-Align & S
        & 0.434 & 0.762 & 0.550 
        & 0.427 & 0.762 & 0.540 
        & 0.411 & 0.772 & 0.530 
        & 0.243 & 0.522 & 0.340 
        & 0.385 & 0.600 & 0.460\\ 

        & BootEA & SS
        & 0.629 & 0.847 & 0.703 
        & 0.622 & 0.853 & 0.701 
        & 0.653 & 0.874 & 0.731 
        & 0.365 & 0.649 & 0.460 
        & 0.594 & 0.818 & 0.666\\ 
        \midrule

        \multirow{7.5}{*}{\rotatebox[origin=c]{90}{\small Structure + Text}}

        & JEANS & S
        & 0.719 & 0.895 & 0.791
        & 0.737 & 0.914 & 0.798 
        & 0.769 & 0.940 & 0.827 
        &- & -&- &- &- &-\\ 
        
        & GM-Align & S
        & 0.679 & 0.785 & - 
        & 0.739 & 0.872 & - 
        & 0.894 & 0.952 & - 
        & 0.574 & 0.646 & 0.602
        & 0.681 & 0.748 & 0.710\\
        
        & HGCN & S
        & 0.720 & 0.857 & 0.760 
        & 0.766 & 0.897 & 0.810 
        & 0.892 & 0.961 & 0.910 
        & 0.670 & 0.770 & 0.710 
        & 0.763 & 0.863 & 0.801\\
        
        & Dual-AMN & SS
        & 0.861 & 0.964 & 0.901 
        & 0.892 & 0.978 & 0.925 
        & 0.954 & 0.994 & 0.970 
        & 0.802 & 0.932 & 0.851 
        & 0.891 & 0.972 & 0.923\\ 
        
        & \seu & U
        & 0.884 & 0.956 & 0.910 
        & 0.947 & 0.984 & 0.961 
        & 0.984 & 0.998 & 0.989 
        & 0.982 & 0.995 & 0.987 
        & 0.983 & 0.994 & 0.987\\

        \cmidrule(l){2-18}
        
        & \textbf{\proposed} & U
        & 0.936 & 0.977 & 0.951
        & 0.972 & 0.994 & 0.979
        & 0.987 & 0.997 & 0.991 
        & 0.990 & 0.994 & 0.991
        & 0.986 & 0.996 & 0.990\\
        
        & \textbf{\proposedpp} & U
        & \textbf{0.947} & \textbf{0.982} & \textbf{0.959}
        & \textbf{0.979} & \textbf{0.996} & \textbf{0.985}
        & {0.991} & {0.998} & {0.993} 
        & \textbf{0.993} & \textbf{0.997} & \textbf{0.995}
        & \textbf{0.989} & \textbf{0.997} & \textbf{0.991} \\ 
        \bottomrule
    \end{tabular}
    }
\end{table*}

%% file: 042validtable.tex
\begin{table}[t]
\caption{The Hit@1 accuracy for entity alignment and relation alignment with different types of textual features.}
    \label{tbl:featvalid}
    \centering
    \small
        \begin{tabular}{cccccc}
            \toprule
            & \multirow{2}{*}{\textbf{Dataset}} & \multirow{2}{*}{\textbf{\glove}} & \textbf{\glove} & \multirow{2}{*}{\textbf{\sbert}} & \textbf{\sbert} \\
            & & & \textbf{+Bigram} & & \textbf{+Bigram} \\
            \midrule
            
            \multirow{5}{*}{\rotatebox[origin=c]{90}{Entity}}
            & {\dbpzh} & 0.813 & 0.909 & 0.920 & 0.929\\
            & {\dbpja} & 0.881 & 0.966 & 0.963 & 0.975\\
            & {\dbpfr} & 0.795 & 0.951 & 0.978 & 0.982\\
            & {\srpfr} & 0.751 & 0.975 & 0.988 & 0.992\\
            & {\srpde} & 0.675 & 0.957 & 0.992 & 0.983\\
            \midrule
            \multirow{5}{*}{\rotatebox[origin=c]{90}{Relation}} 
            & {\dbpzh} & 0.510 & 0.942 & 0.974 & 0.997\\
            & {\dbpja} & 0.602 & 0.974 & 0.983 & 0.990\\
            & {\dbpfr} & 0.640 & 0.912 & 0.954 & 0.973\\
            & {\srpfr} & 0.716 & 0.980 & 0.980 & 0.990\\
            & {\srpde} & 0.704 & 1.000 & 0.969 & 1.000\\
            \bottomrule
        \end{tabular}
\end{table}

%% file: 045dropratio.tex
\begingroup
\begin{table}[t]
\renewcommand{\arraystretch}{0.6}
\caption{The Hit@1 accuracy for varying percentages of dropped triples in the target graph: 0\%, 25\%, 50\%, and 75\%.}
    \label{tbl:dropratio}
    \centering
    \resizebox{0.9\linewidth}{!}{
    \begin{tabular}{cccc}
        \toprule[0.4pt]
        {\textscale{0.625 }{\textbf{Drop}}} & {\textscale{0.62 }{\textbf{Initialization}}} & {\textscale{0.62 }{\textbf{Refinement}}} & {\textscale{0.62 }{\textbf{Verification}}} \\
        {\textscale{0.625 }{\textbf{Ratio}}} & {\textscale{0.625 }{\textbf{Step 1}}} & {\textscale{0.625 }{\textbf{Step 2}}} & {\textscale{0.625 }{\textbf{Step 3}}}\\
        \midrule[0.01pt]

        {\textscale{0.628 }{\textbf{0\%}}}  & {\textscale{0.628 }{0.929}} & {\textscale{0.628 }{0.943 (+0.014)}} & {\textscale{0.628 }{0.946 (+0.003)}} \\
        {\textscale{0.628 }{\textbf{25\%}}} & {\textscale{0.628 }{0.920}} & {\textscale{0.628 }{0.931 (+0.011)}} & {\textscale{0.628 }{0.934 (+0.003)}} \\
        {\textscale{0.628 }{\textbf{50\%}}} & {\textscale{0.628 }{0.899}} & {\textscale{0.628 }{0.906 (+0.007)}} & {\textscale{0.628 }{0.910 (+0.004)}} \\
        {\textscale{0.628 }{\textbf{75\%}}} & {\textscale{0.628 }{0.860}} & {\textscale{0.628 }{0.854 (-0.006)}} & {\textscale{0.628 }{0.862 (+0.008)}} \\
       \bottomrule[0.4pt]
    \end{tabular}
    }
\end{table}
\endgroup

%% file: 043casestudy.tex
\begin{CJK*}{UTF8}{gbsn}

\begin{table*}[t]
\caption{An example of erroneous alignment correction, Dataset: \dbpzh. For the source entity ``赣语''($\rightarrow$ ``Gan language''), our correction step raises the rank of the correct target entity ``Gan Chinese'' ($\rightarrow$ ``Gan Chinese'') from 7 to 1, while it lowers the rank of the wrong target entity ``Kam language'' ($\rightarrow$ ``Kam language'' ) from 1 to 9.}
    \label{tbl:casestudy}
    \centering
    \resizebox{0.99\linewidth}{!}{%
    \begin{tabular}{cQQQ}
        \toprule
        
        & \multicolumn{1}{c}{\textbf{(Source) Entity}} & \multicolumn{1}{c}{\textbf{(Target) Correct Entity}} & \multicolumn{1}{c}{\textbf{(Target) Wrong Entity}}  \\ 
        \cmidrule(lr){1-1}\cmidrule(lr){2-2}\cmidrule(lr){3-3}\cmidrule(lr){4-4}
 
        
        \multirow{2}{*}{\makecell{\textbf{Name} (\textit{entity})}}
        & ``赣语'' & ``Gan Chinese'' & ``Kam language'' \\
        & $\rightarrow$ ``Gan language'' & $\rightarrow$ ``Gan Chinese'' & $\rightarrow$ ``Kam language'' \\
        \cmidrule(lr){1-1}\cmidrule(lr){2-2}\cmidrule(lr){3-3}\cmidrule(lr){4-4}
        
        \multirow{9.5}{*}{\makecell{\textbf{Neighbor Triples} \\ (\textit{relation}, \textit{entity})}}

        & (``fam'', ``汉藏语系'') & (``fam'', ``Sinitic languages'')	&  (``langs'', ``Kam people'') \\
        & $\rightarrow$ (``FAM'', ``Sino-Tibetan'') & $\rightarrow$ (``fam'', ``Sinitic languages'')	&  $\rightarrow$ (``langs'', ``Kam people'') \\
        \cmidrule(lr){2-2}\cmidrule(lr){3-3}\cmidrule(lr){4-4}

        & (``南'', ``客家语'') & (``translit Lang 1 Type'', ``Jiangxi'') & (``region'', ``Guangxi'') \\
        & $\rightarrow$ (``South'', ``Hakka'') & $\rightarrow$ (``translit Lang 1 Type'', ``Jiangxi'') & $\rightarrow$ (``region'', ``Guangxi'') \\
        \cmidrule(lr){2-2}\cmidrule(lr){3-3}\cmidrule(lr){4-4}

        & (``地区'', ``安徽省'') & (``states'', ``China'') & (``region'', ``Guizhou'')  \\
        & $\rightarrow$ (``region'', ``Anhui Province'') & $\rightarrow$ (``states'', ``China'') & $\rightarrow$ (``region'', ``Guizhou'') \\
        \cmidrule(lr){2-2}\cmidrule(lr){3-3}\cmidrule(lr){4-4}

        & (``fam'', ``汉语族'') & (``fam'', ``Chinese language'') & (``region'', ``Hunan'')  \\
        & $\rightarrow$ (``FAM'', ``Chinese'') & $\rightarrow$ (``fam'', ``Chinese language'') & $\rightarrow$ (``region'', ``Hunan'') \\

        \bottomrule
    \end{tabular}
    }
\end{table*}

\end{CJK*}

%% file: 044ablation.tex
\begin{table}[t]
\caption{Ablation analysis on each component of \proposed.}
    \label{tbl:ablation}
    \centering
    \resizebox{0.99\linewidth}{!}{%
    \begin{tabular}{lZZZZZZ}
        \toprule
        
        \multirow{2.5}{*}{\textbf{Method}}
        & \multicolumn{3}{c}{\textbf{\dbpzh}} 
        & \multicolumn{3}{c}{\textbf{\dbpja}} \\ 
        \cmidrule(lr){2-4}\cmidrule(lr){5-7}

        & \small H@1 & \small H@10 & \small MRR
        & \small H@1 & \small H@10 & \small MRR\\ 
        \midrule

        \textbf{\proposed}
        & \textbf{0.936} & \textbf{0.977} & \textbf{0.951}
        & \textbf{0.972} & \textbf{0.994} & \textbf{0.979} \\
        \midrule

        \ w/o \sbert ($\rightarrow$ \glove)
        & 0.910 & 0.964 & 0.930 
        & 0.967 & 0.992 & 0.976 \\ 

        \ w/o Bigram
        & 0.934 & 0.975 & 0.949 
        & 0.971 & 0.993 & 0.979 \\ 
        \midrule

        \textbf{\proposedpp}
        & \textbf{0.947} & \textbf{0.982} & \textbf{0.959} 
        & \textbf{0.979} & \textbf{0.996} & \textbf{0.985} \\
        \midrule

        \ w/o Conf
        & 0.938 & 0.978 & 0.952 
        & 0.973 & 0.994 & 0.981 \\  

        \ w/o Cons
        & 0.941 & 0.981 & 0.957 
        & 0.976 & 0.994 & 0.983 \\  
        \midrule


        \ w/o Triple sorting
        & 0.928 & 0.974 & 0.949 
        & 0.965 & 0.990 & 0.972 \\  

        \ w/o STS ($\rightarrow$ NLI)
        & 0.943 & 0.976 & 0.954
        & 0.975 & 0.996 & 0.982 \\ 
        
       \bottomrule
    \end{tabular}
    }
\end{table}

%% file: 050conclusion.tex
\proposed is a novel unsupervised cross-lingual EA pipeline that effectively leverages semantic textual features of both entities and relations.
Its robust alignment process consists of three key steps: 
(1) obtaining alignment scores of entities and relations using both original and dual KGs,
(2) iteratively refining these scores by fusing entity-level and relation-level alignment via neighbor triple matching, and 
(3) verifying erroneous alignments by examining neighbor triples as linearized texts.
\proposed effectively enhances the base methods used for initial alignment scoring
and provides a flexible framework adaptable to various EA approaches.

Our experiments demonstrate that \proposed achieves superior performance and strong robustness to noisy textual features. 
It also performs effectively even in non-isomorphic scenarios, unlike many other methods that assume isomorphism. 
The \textit{Align-then-Verify} pipeline of \proposed shows general applicability, consistently improving the performance of base EA methods.
Furthermore, the modular design of \proposed ensures compatibility with future state-of-the-art methods, offering opportunities for future performance gains as new techniques emerge.
These findings emphasize the importance of combining semantic and lexical features at the triple level, while employing iterative refinement and verification steps, to achieve high accuracy in cross-lingual EA tasks.

%% file: 060appendix.tex
\section{Template for Triple Linearization}
\label{sec:template}
To linearize an entity's relation triples, we employ a straightforward text template that succinctly summarizes relevant knowledge, focusing on the entity.
\begin{equation}
\begin{split}
    Z_S(i) &= \text{``}[\texttt{Entity} \text{ } i], \text{which ''} \\
    &+ \Vert_{(i, p, i')\in\mathcal{T}'} \text{``}[\texttt{Relation} \text{ } p] \text{ is } [\texttt{Entity} \text{ } i']\text{, ''}
\end{split}
\end{equation}
For instance, in the DBP\textsubscript{EN} KG, the entity ``Football League One'' can be represented through linearized text based on its neighbor triples:
``Football League One, which relegation is Football League Championship, promotion is Football League Championship, promotion is Football League Two, relegation is Football League Two, league is Sheffield United F.C..''.
An alternative approach to linearization involves employing language models fine-tuned to transform a subgraph (e.g., 1-hop triple knowledge) into plausible text~\cite{graph2text}.
This avenue is left for future exploration.

\section{Erroneous Alignment Verification}
\label{sec:verfmatrix}
The efficacy of the verification step is analyzed in detail by examining the verification matrix in Table~\ref{tbl:verfmat}, which summarizes the counts of correctly/incorrectly aligned entity pairs before and after the correction process.
Entities labeled as ``non-detected'' are excluded from correction (i.e., $\notin\mathcal{E}^{ver}$), and entities classified as ``non-corrected'' do not pass the cross-verification, resulting in no adjustments to the ranked list.
For both non-detected and non-corrected entities, where no modifications were made, most cases are prove difficult to rectify solely based on the semantic textual relevance of entities' neighbor triples;
this challenge mostly arises from structural disparities between the source and target KGs.
In the case of corrected alignments, over 90\% of corrections result in accurate alignments (i.e., $\bm{\times\rightarrow\checkmark}$ and $\bm{\checkmark\rightarrow\checkmark}$), significantly improving overall alignment accuracy.
Furthermore, the effectiveness of our verification step becomes increasingly evident as noise levels increase, strongly indicating that the additional verification step greatly enhances alignment accuracy under textual noise.

\begin{table}[t]
\caption{Verification matrices illustrating the number of corrected erroneous alignments by \proposedpp, Dataset: \dbpzh. Entities labeled as ``non-detected'' are not considered for correction (i.e., $\notin\mathcal{E}^{ver}$), and entities classified as ``non-corrected'' fail the cross-verification, resulting in no adjustments to the ranked list.}
    \label{tbl:verfmat}
    \centering
    \resizebox{0.99\linewidth}{!}{
        \begin{tabular}{ccPPPP}
            \toprule
            & & \multicolumn{3}{c}{\textbf{Detected}} & {\textbf{Non-Detected}} \\\cmidrule(lr){3-5}\cmidrule(lr){6-6}
            & & \multicolumn{2}{c}{\textbf{Before Correction}} & \multirow{2}{*}{\makecell{Non-\\Corrected}} & \multirow{2}{*}{\makecell{Non-\\Detected}} \\
            & & Wrong & Correct & & \\
            \midrule
            
            \multirow{6.5}{*}{\rotatebox[origin=c]{90}{\textbf{Noise 0\%}}} & \multirow{3}{*}{\rotatebox[origin=c]{90}{{Wrong}}} 

            & $\bm{\times \rightarrow \times}$ & $\bm{\checkmark \rightarrow \times}$ & $\bm{\times (\rightarrow \times)}$ & $\bm{\times (\rightarrow \times)}$  \\
            & & 26 & 43 & 505 & 262 \\
            & & (2.9\%) & (4.7\%) & &  \\
            \cmidrule(l){2-6}
            
            & \multirow{3}{*}{\rotatebox[origin=c]{90}{{Correct}}} 
            
            & $\bm{\times \rightarrow \checkmark}$ & $\bm{\checkmark \rightarrow \checkmark}$ & $\bm{\checkmark (\rightarrow \checkmark)}$ & $\bm{\checkmark (\rightarrow \checkmark)}$ \\
            & & 88 & 753 & 1,494 & 11,910 \\
            & & (9.7\%) & (82.7\%) & &  \\
            
            \midrule
            \multirow{6.5}{*}{\rotatebox[origin=c]{90}{\textbf{Noise 10\%}}} & \multirow{3}{*}{\rotatebox[origin=c]{90}{{Wrong}}} 

            & $\bm{\times \rightarrow \times}$ & $\bm{\checkmark \rightarrow \times}$ & $\bm{\times (\rightarrow \times)}$ & $\bm{\times (\rightarrow \times)}$  \\
            & & 33 & 44 & 534 & 370 \\
            & & (2.9\%) & (3.8\%) & &  \\
            \cmidrule(l){2-6}
            
            & \multirow{3}{*}{\rotatebox[origin=c]{90}{{Correct}}} 
            
            & $\bm{\times \rightarrow \checkmark}$ & $\bm{\checkmark \rightarrow \checkmark}$ & $\bm{\checkmark (\rightarrow \checkmark)}$ & $\bm{\checkmark (\rightarrow \checkmark)}$ \\
            & & 180 & 894 & 1,149 & 11,796 \\
            & & (15.6\%) & (77.7\%) & &  \\
            
            \midrule
            
            \multirow{6.5}{*}{\rotatebox[origin=c]{90}{\textbf{Noise 20\%}}} & \multirow{3}{*}{\rotatebox[origin=c]{90}{{Wrong}}} 

            & $\bm{\times \rightarrow \times}$ & $\bm{\checkmark \rightarrow \times}$ & $\bm{\times (\rightarrow \times)}$ & $\bm{\times (\rightarrow \times)}$  \\
            & & 38 & 48 & 568 & 512 \\
            & & (2.8\%) & (3.5\%) & &  \\
            \cmidrule(l){2-6}
            
            & \multirow{3}{*}{\rotatebox[origin=c]{90}{{Correct}}} 
            
            & $\bm{\times \rightarrow \checkmark}$ & $\bm{\checkmark \rightarrow \checkmark}$ & $\bm{\checkmark (\rightarrow \checkmark)}$ & $\bm{\checkmark (\rightarrow \checkmark)}$ \\
            & & 256 & 1,026 & 882 & 11,670 \\
            & & (18.7\%) & (75.0\%) & &  \\
            \bottomrule
        \end{tabular}
    }
\end{table}

\section{Textual Noise Simulation in KGs}
\label{sec:noise}
The robustness of \proposed and \proposedpp under textual feature noise (discussed in Section~\ref{subsubsec:noise}) is assessed through experiments simulating various types of text translation errors that commonly observed as noise. 
The three distinct categories of textual noise are outlined as follows:
\begin{itemize}
    \item \textbf{Phonetic errors}: 
    These errors involve variations in the representation of words based on their phonetic similarity. 
    For instance, substitutions like
    `intu'/`into', `o'/`oe', `li'/`ri', `ty'/`ti', `cu'/`ka', `ca'/`ka', `ar'/`al', `tic'/`th', `se'/`th', `nes'/`nais', `ud'/`ade', `Ji'/`Gi', `fi'/`fy', `ps'/`pus', `er'/`ar', `our'/`ur', `ar'/`ur', `la'/`ra', `ei'/`ee', `ny'/`ni', `ew'/`ou', `ar'/`or', `or'/`ol', `ol'/`oul', `ry'/`ly', `wi'/`wy', or `ic'/`ik'
    are considered phonetic errors.

    \item \textbf{Missing characters}: 
    This type of error occurs when certain characters are omitted from a word. 
    Examples include instances where `-s', `-e', or ` ' (blank) are missing.
    
    \item \textbf{Attached characters}:
    Attached characters refer to errors where additional characters are affixed to a word. 
    For example, errors such as `-s' or `-e' being added to a word fall into this category.
\end{itemize}

\section{Compatibility of \proposed with \eva}
\label{sec:application}
The compatibility of \proposed (Section~\ref{subsubsec:application}) is explored using entity-level and relation-level alignment scores initially obtained by \eva~\cite{eva}, a self-supervised cross-lingual EA method.
Originally designed for unsupervised multi-modality entity alignment (MMEA), \eva leverages visual similarity among entities to generate a pseudo seed alignments (i.e., visual pivot dictionaries) for model training. 
Since this study only focuses on textual and structural information, \eva is tailored to use textual information (i.e., entity names for original KGs and relation names for dual KGs) in place of visual information to collect pseudo seed alignments. 
Final textual embeddings are obtained by concatenating \sbert (as semantic features) and Bigram (as lexical features).

For initial seeds, we collect top-$k$ entity pairs based on the similarity of their textual embeddings.
To keep the initial seeds small, $k$ is set to 10\% of the total number of ground-truth pairs.
Specifically, the number of initial seeds for entity alignment is 1500, while for relation alignment, it is 89 (\dbpzh) and 58 (\dbpja).

We use 2-layer graph convolution networks (GCNs) with input, hidden, and output dimension of 128.
As the number of initial seeds are small, \eva adopts the iterative learning (IL) strategy. 
IL expands the training set by updating the candidate set during training and merging it into the training set at certain epochs.
The training process runs for 1000 epochs, with IL applied after the first 500 epochs.
The candidate set is updated every 10 epochs and merged into training set every 10 updates (i.e., 100 epochs).

For optimization, AdamW~\cite{adamw} is used as the optimizer with a learning rate of $5e-4$ and a weight decay of $1e-2$.
The output alignment scores from \eva serve as the initial alignment scores (Step 1) within our \proposed framework in Section~\ref{subsubsec:application}.
Table~\ref{tbl:eva} reports the accuracy of \eva in terms of both entity-level and relation-level alignment between the source and target KGs.

\begin{table}[t]
\caption{The accuracy of initial alignment scores}
    \label{tbl:eva}
    \centering
    \resizebox{0.99\linewidth}{!}{%
    \begin{tabular}{cccccccc}
        \toprule
        
        \multicolumn{2}{c}{\multirow{2.5}{*}{\textbf{Alignment Score}}}
        & \multicolumn{3}{c}{\textbf{\dbpzh}} 
        & \multicolumn{3}{c}{\textbf{\dbpja}} \\ 
        \cmidrule(lr){3-5}\cmidrule(lr){6-8}

        & 
        & \small H@1 & \small H@10 & \small MRR
        & \small H@1 & \small H@10 & \small MRR\\ 
        \midrule

        \multirow{2}{*}{\rotatebox[origin=c]{0}{\small \eva}}
        & Entity-level
        & {0.828} & {0.923} & {0.864}
        & {0.917} & {0.974} & {0.979} \\
        & Relation-level
        & {0.965} & {0.997} & {0.980}
        & {0.935} & {1.000} & {0.966} \\
        \midrule

        \multirow{2}{*}{\rotatebox[origin=c]{0}{\small \seu}}
        & Entity-level
        & {0.929} & {0.972} & {0.945}
        & {0.975} & {0.994} & {0.982} \\
        & Relation-level
        & {0.992} & {1.000} & {0.996}
        & {0.983} & {1.000} & {0.990} \\

       \bottomrule
    \end{tabular}
    }
\end{table}

%% file: main.bbl

\begin{thebibliography}{47}


\ifx \showCODEN    \undefined \def \showCODEN     #1{\unskip}     \fi
\ifx \showDOI      \undefined \def \showDOI       #1{#1}\fi
\ifx \showISBNx    \undefined \def \showISBNx     #1{\unskip}     \fi
\ifx \showISBNxiii \undefined \def \showISBNxiii  #1{\unskip}     \fi
\ifx \showISSN     \undefined \def \showISSN      #1{\unskip}     \fi
\ifx \showLCCN     \undefined \def \showLCCN      #1{\unskip}     \fi
\ifx \shownote     \undefined \def \shownote      #1{#1}          \fi
\ifx \showarticletitle \undefined \def \showarticletitle #1{#1}   \fi
\ifx \showURL      \undefined \def \showURL       {\relax}        \fi
\providecommand\bibfield[2]{#2}
\providecommand\bibinfo[2]{#2}
\providecommand\natexlab[1]{#1}
\providecommand\showeprint[2][]{arXiv:#2}

\bibitem[Auer et~al\mbox{.}(2007)]%
        {kgs_dbpdia}
\bibfield{author}{\bibinfo{person}{S{\"o}ren Auer}, \bibinfo{person}{Christian Bizer}, \bibinfo{person}{Georgi Kobilarov}, \bibinfo{person}{Jens Lehmann}, \bibinfo{person}{Richard Cyganiak}, {and} \bibinfo{person}{Zachary Ives}.} \bibinfo{year}{2007}\natexlab{}.
\newblock \showarticletitle{DBpedia: A Nucleus for a Web of Open Data}. In \bibinfo{booktitle}{\emph{The Semantic Web}}, \bibfield{editor}{\bibinfo{person}{Karl Aberer}, \bibinfo{person}{Key-Sun Choi}, \bibinfo{person}{Natasha Noy}, \bibinfo{person}{Dean Allemang}, \bibinfo{person}{Kyung-Il Lee}, \bibinfo{person}{Lyndon Nixon}, \bibinfo{person}{Jennifer Golbeck}, \bibinfo{person}{Peter Mika}, \bibinfo{person}{Diana Maynard}, \bibinfo{person}{Riichiro Mizoguchi}, \bibinfo{person}{Guus Schreiber}, {and} \bibinfo{person}{Philippe Cudr{\'e}-Mauroux}} (Eds.). \bibinfo{publisher}{Springer Berlin Heidelberg}, \bibinfo{address}{Berlin, Heidelberg}, \bibinfo{pages}{722--735}.
\newblock
\showISBNx{978-3-540-76298-0}


\bibitem[Bordes et~al\mbox{.}(2013)]%
        {transe}
\bibfield{author}{\bibinfo{person}{Antoine Bordes}, \bibinfo{person}{Nicolas Usunier}, \bibinfo{person}{Alberto Garcia-Duran}, \bibinfo{person}{Jason Weston}, {and} \bibinfo{person}{Oksana Yakhnenko}.} \bibinfo{year}{2013}\natexlab{}.
\newblock \showarticletitle{Translating Embeddings for Modeling Multi-relational Data}. In \bibinfo{booktitle}{\emph{Advances in Neural Information Processing Systems}}, \bibfield{editor}{\bibinfo{person}{C.J. Burges}, \bibinfo{person}{L.~Bottou}, \bibinfo{person}{M.~Welling}, \bibinfo{person}{Z.~Ghahramani}, {and} \bibinfo{person}{K.Q. Weinberger}} (Eds.), Vol.~\bibinfo{volume}{26}. \bibinfo{publisher}{Curran Associates, Inc.}
\newblock
\urldef\tempurl%
\url{https://proceedings.neurips.cc/paper_files/paper/2013/file/1cecc7a77928ca8133fa24680a88d2f9-Paper.pdf}
\showURL{%
\tempurl}


\bibitem[Carlson et~al\mbox{.}(2010)]%
        {kgs_nell}
\bibfield{author}{\bibinfo{person}{Andrew Carlson}, \bibinfo{person}{Justin Betteridge}, \bibinfo{person}{Bryan Kisiel}, \bibinfo{person}{Burr Settles}, \bibinfo{person}{Estevam~R. Hruschka}, {and} \bibinfo{person}{Tom~M. Mitchell}.} \bibinfo{year}{2010}\natexlab{}.
\newblock \showarticletitle{Toward an Architecture for Never-Ending Language Learning}. In \bibinfo{booktitle}{\emph{Proceedings of the Twenty-Fourth AAAI Conference on Artificial Intelligence}} (Atlanta, Georgia) \emph{(\bibinfo{series}{AAAI'10})}. \bibinfo{publisher}{AAAI Press}, \bibinfo{pages}{1306–1313}.
\newblock


\bibitem[Chen et~al\mbox{.}(2021)]%
        {jeans}
\bibfield{author}{\bibinfo{person}{Muhao Chen}, \bibinfo{person}{Weijia Shi}, \bibinfo{person}{Ben Zhou}, {and} \bibinfo{person}{Dan Roth}.} \bibinfo{year}{2021}\natexlab{}.
\newblock \showarticletitle{Cross-lingual Entity Alignment with Incidental Supervision}. In \bibinfo{booktitle}{\emph{Proceedings of the 16th Conference of the European Chapter of the Association for Computational Linguistics: Main Volume}}. \bibinfo{publisher}{Association for Computational Linguistics}, \bibinfo{address}{Online}, \bibinfo{pages}{645--658}.
\newblock
\urldef\tempurl%
\url{https://doi.org/10.18653/v1/2021.eacl-main.53}
\showDOI{\tempurl}


\bibitem[Chen et~al\mbox{.}(2018)]%
        {KDCoE}
\bibfield{author}{\bibinfo{person}{Muhao Chen}, \bibinfo{person}{Yingtao Tian}, \bibinfo{person}{Kai-Wei Chang}, \bibinfo{person}{Steven Skiena}, {and} \bibinfo{person}{Carlo Zaniolo}.} \bibinfo{year}{2018}\natexlab{}.
\newblock \showarticletitle{Co-Training Embeddings of Knowledge Graphs and Entity Descriptions for Cross-Lingual Entity Alignment}. In \bibinfo{booktitle}{\emph{Proceedings of the 27th International Joint Conference on Artificial Intelligence}} (Stockholm, Sweden) \emph{(\bibinfo{series}{IJCAI'18})}. \bibinfo{publisher}{AAAI Press}, \bibinfo{pages}{3998–4004}.
\newblock
\showISBNx{9780999241127}


\bibitem[Chen et~al\mbox{.}(2017)]%
        {mtranse}
\bibfield{author}{\bibinfo{person}{Muhao Chen}, \bibinfo{person}{Yingtao Tian}, \bibinfo{person}{Mohan Yang}, {and} \bibinfo{person}{Carlo Zaniolo}.} \bibinfo{year}{2017}\natexlab{}.
\newblock \showarticletitle{Multilingual Knowledge Graph Embeddings for Cross-Lingual Knowledge Alignment}. In \bibinfo{booktitle}{\emph{Proceedings of the 26th International Joint Conference on Artificial Intelligence}} (Melbourne, Australia) \emph{(\bibinfo{series}{IJCAI'17})}. \bibinfo{publisher}{AAAI Press}, \bibinfo{pages}{1511–1517}.
\newblock
\showISBNx{9780999241103}


\bibitem[Chen et~al\mbox{.}(2023)]%
        {meaformer}
\bibfield{author}{\bibinfo{person}{Zhuo Chen}, \bibinfo{person}{Jiaoyan Chen}, \bibinfo{person}{Wen Zhang}, \bibinfo{person}{Lingbing Guo}, \bibinfo{person}{Yin Fang}, \bibinfo{person}{Yufeng Huang}, \bibinfo{person}{Yichi Zhang}, \bibinfo{person}{Yuxia Geng}, \bibinfo{person}{Jeff~Z. Pan}, \bibinfo{person}{Wenting Song}, {and} \bibinfo{person}{Huajun Chen}.} \bibinfo{year}{2023}\natexlab{}.
\newblock \showarticletitle{MEAformer: Multi-modal Entity Alignment Transformer for Meta Modality Hybrid}. In \bibinfo{booktitle}{\emph{{ACM} Multimedia}}. \bibinfo{publisher}{{ACM}}.
\newblock


\bibitem[Cuturi(2013)]%
        {sinkhorn}
\bibfield{author}{\bibinfo{person}{Marco Cuturi}.} \bibinfo{year}{2013}\natexlab{}.
\newblock \showarticletitle{Sinkhorn Distances: Lightspeed Computation of Optimal Transport}. In \bibinfo{booktitle}{\emph{Advances in Neural Information Processing Systems}}, \bibfield{editor}{\bibinfo{person}{C.J. Burges}, \bibinfo{person}{L.~Bottou}, \bibinfo{person}{M.~Welling}, \bibinfo{person}{Z.~Ghahramani}, {and} \bibinfo{person}{K.Q. Weinberger}} (Eds.), Vol.~\bibinfo{volume}{26}. \bibinfo{publisher}{Curran Associates, Inc.}
\newblock
\urldef\tempurl%
\url{https://proceedings.neurips.cc/paper_files/paper/2013/file/af21d0c97db2e27e13572cbf59eb343d-Paper.pdf}
\showURL{%
\tempurl}


\bibitem[Devlin et~al\mbox{.}(2018)]%
        {bert}
\bibfield{author}{\bibinfo{person}{Jacob Devlin}, \bibinfo{person}{Ming-Wei Chang}, \bibinfo{person}{Kenton Lee}, {and} \bibinfo{person}{Kristina Toutanova}.} \bibinfo{year}{2018}\natexlab{}.
\newblock \showarticletitle{Bert: Pre-training of deep bidirectional transformers for language understanding}.
\newblock \bibinfo{journal}{\emph{arXiv preprint arXiv:1810.04805}} (\bibinfo{year}{2018}).
\newblock


\bibitem[Ge et~al\mbox{.}(2021)]%
        {EASY}
\bibfield{author}{\bibinfo{person}{Congcong Ge}, \bibinfo{person}{Xiaoze Liu}, \bibinfo{person}{Lu Chen}, \bibinfo{person}{Baihua Zheng}, {and} \bibinfo{person}{Yunjun Gao}.} \bibinfo{year}{2021}\natexlab{}.
\newblock \showarticletitle{Make it easy: An effective end-to-end entity alignment framework}. In \bibinfo{booktitle}{\emph{Proceedings of the 44th International ACM SIGIR Conference on Research and Development in Information Retrieval}}. \bibinfo{pages}{777--786}.
\newblock


\bibitem[Guo et~al\mbox{.}(2019)]%
        {dataset_srprs}
\bibfield{author}{\bibinfo{person}{Lingbing Guo}, \bibinfo{person}{Zequn Sun}, {and} \bibinfo{person}{Wei Hu}.} \bibinfo{year}{2019}\natexlab{}.
\newblock \showarticletitle{Learning to Exploit Long-term Relational Dependencies in Knowledge Graphs}. In \bibinfo{booktitle}{\emph{Proceedings of the 36th International Conference on Machine Learning}} \emph{(\bibinfo{series}{Proceedings of Machine Learning Research}, Vol.~\bibinfo{volume}{97})}, \bibfield{editor}{\bibinfo{person}{Kamalika Chaudhuri} {and} \bibinfo{person}{Ruslan Salakhutdinov}} (Eds.). \bibinfo{publisher}{PMLR}, \bibinfo{pages}{2505--2514}.
\newblock
\urldef\tempurl%
\url{https://proceedings.mlr.press/v97/guo19c.html}
\showURL{%
\tempurl}


\bibitem[Han et~al\mbox{.}(2020)]%
        {qa2}
\bibfield{author}{\bibinfo{person}{Jiale Han}, \bibinfo{person}{Bo Cheng}, {and} \bibinfo{person}{Xu Wang}.} \bibinfo{year}{2020}\natexlab{}.
\newblock \showarticletitle{Open Domain Question Answering based on Text Enhanced Knowledge Graph with Hyperedge Infusion}. In \bibinfo{booktitle}{\emph{Findings}}.
\newblock
\urldef\tempurl%
\url{https://aclanthology.org/2020.findings-emnlp.133.pdf}
\showURL{%
\tempurl}


\bibitem[Jiang et~al\mbox{.}(2023)]%
        {Jiang2023UnsupervisedDC}
\bibfield{author}{\bibinfo{person}{Chuanyu Jiang}, \bibinfo{person}{Yiming Qian}, \bibinfo{person}{Lijun Chen}, \bibinfo{person}{Yang Gu}, {and} \bibinfo{person}{Xia Xie}.} \bibinfo{year}{2023}\natexlab{}.
\newblock \showarticletitle{Unsupervised Deep Cross-Language Entity Alignment}.
\newblock \bibinfo{journal}{\emph{ArXiv}}  \bibinfo{volume}{abs/2309.10598} (\bibinfo{year}{2023}).
\newblock


\bibitem[Kuhn(1955)]%
        {hungarian}
\bibfield{author}{\bibinfo{person}{Harold~W. Kuhn}.} \bibinfo{year}{1955}\natexlab{}.
\newblock \showarticletitle{The Hungarian method for the assignment problem}.
\newblock \bibinfo{journal}{\emph{Naval Research Logistics (NRL)}}  \bibinfo{volume}{52} (\bibinfo{year}{1955}).
\newblock
\urldef\tempurl%
\url{https://api.semanticscholar.org/CorpusID:9426884}
\showURL{%
\tempurl}


\bibitem[Lin et~al\mbox{.}(2019)]%
        {cr1}
\bibfield{author}{\bibinfo{person}{Bill~Yuchen Lin}, \bibinfo{person}{Xinyue Chen}, \bibinfo{person}{Jamin Chen}, {and} \bibinfo{person}{Xiang Ren}.} \bibinfo{year}{2019}\natexlab{}.
\newblock \showarticletitle{KagNet: Knowledge-Aware Graph Networks for Commonsense Reasoning}. In \bibinfo{booktitle}{\emph{Proceedings of the 2019 Conference on Empirical Methods in Natural Language Processing and the 9th International Joint Conference on Natural Language Processing (EMNLP-IJCNLP)}}. \bibinfo{pages}{2829--2839}.
\newblock


\bibitem[Liu et~al\mbox{.}(2020b)]%
        {eva}
\bibfield{author}{\bibinfo{person}{Fangyu Liu}, \bibinfo{person}{Muhao Chen}, \bibinfo{person}{Dan Roth}, {and} \bibinfo{person}{Nigel Collier}.} \bibinfo{year}{2020}\natexlab{b}.
\newblock \showarticletitle{Visual Pivoting for (Unsupervised) Entity Alignment}.
\newblock \bibinfo{journal}{\emph{ArXiv}}  \bibinfo{volume}{abs/2009.13603} (\bibinfo{year}{2020}).
\newblock
\urldef\tempurl%
\url{https://api.semanticscholar.org/CorpusID:221995513}
\showURL{%
\tempurl}


\bibitem[Liu et~al\mbox{.}(2023)]%
        {dualmatch}
\bibfield{author}{\bibinfo{person}{Xiaoze Liu}, \bibinfo{person}{Junyang Wu}, \bibinfo{person}{Tianyi Li}, \bibinfo{person}{Lu Chen}, {and} \bibinfo{person}{Yunjun Gao}.} \bibinfo{year}{2023}\natexlab{}.
\newblock \showarticletitle{Unsupervised Entity Alignment for Temporal Knowledge Graphs}.
\newblock \bibinfo{journal}{\emph{Proceedings of the ACM Web Conference 2023}} (\bibinfo{year}{2023}).
\newblock
\urldef\tempurl%
\url{https://api.semanticscholar.org/CorpusID:256503992}
\showURL{%
\tempurl}


\bibitem[Liu et~al\mbox{.}(2019)]%
        {roberta}
\bibfield{author}{\bibinfo{person}{Yinhan Liu}, \bibinfo{person}{Myle Ott}, \bibinfo{person}{Naman Goyal}, \bibinfo{person}{Jingfei Du}, \bibinfo{person}{Mandar Joshi}, \bibinfo{person}{Danqi Chen}, \bibinfo{person}{Omer Levy}, \bibinfo{person}{Mike Lewis}, \bibinfo{person}{Luke Zettlemoyer}, {and} \bibinfo{person}{Veselin Stoyanov}.} \bibinfo{year}{2019}\natexlab{}.
\newblock \showarticletitle{Roberta: A robustly optimized bert pretraining approach}.
\newblock \bibinfo{journal}{\emph{arXiv preprint arXiv:1907.11692}} (\bibinfo{year}{2019}).
\newblock


\bibitem[Liu et~al\mbox{.}(2021)]%
        {cr2}
\bibfield{author}{\bibinfo{person}{Ye Liu}, \bibinfo{person}{Yao Wan}, \bibinfo{person}{Lifang He}, \bibinfo{person}{Hao Peng}, {and} \bibinfo{person}{S~Yu Philip}.} \bibinfo{year}{2021}\natexlab{}.
\newblock \showarticletitle{Kg-bart: Knowledge graph-augmented bart for generative commonsense reasoning}. In \bibinfo{booktitle}{\emph{Proceedings of the AAAI Conference on Artificial Intelligence}}, Vol.~\bibinfo{volume}{35}. \bibinfo{pages}{6418--6425}.
\newblock


\bibitem[Liu et~al\mbox{.}(2020a)]%
        {AttrGNN}
\bibfield{author}{\bibinfo{person}{Zhiyuan Liu}, \bibinfo{person}{Yixin Cao}, \bibinfo{person}{Liangming Pan}, \bibinfo{person}{Juanzi Li}, {and} \bibinfo{person}{Tat-Seng Chua}.} \bibinfo{year}{2020}\natexlab{a}.
\newblock \showarticletitle{Exploring and Evaluating Attributes, Values, and Structures for Entity Alignment}. \bibinfo{pages}{6355--6364}.
\newblock
\urldef\tempurl%
\url{https://doi.org/10.18653/v1/2020.emnlp-main.515}
\showDOI{\tempurl}


\bibitem[Loshchilov and Hutter(2017)]%
        {adamw}
\bibfield{author}{\bibinfo{person}{Ilya Loshchilov} {and} \bibinfo{person}{Frank Hutter}.} \bibinfo{year}{2017}\natexlab{}.
\newblock \showarticletitle{Decoupled Weight Decay Regularization}. In \bibinfo{booktitle}{\emph{International Conference on Learning Representations}}.
\newblock
\urldef\tempurl%
\url{https://api.semanticscholar.org/CorpusID:53592270}
\showURL{%
\tempurl}


\bibitem[Mao et~al\mbox{.}(2021a)]%
        {DualAMN}
\bibfield{author}{\bibinfo{person}{Xin Mao}, \bibinfo{person}{Wenting Wang}, \bibinfo{person}{Yuanbin Wu}, {and} \bibinfo{person}{Man Lan}.} \bibinfo{year}{2021}\natexlab{a}.
\newblock \showarticletitle{Boosting the speed of entity alignment 10$\times$: Dual attention matching network with normalized hard sample mining}. In \bibinfo{booktitle}{\emph{Proceedings of the Web Conference 2021}}. \bibinfo{pages}{821--832}.
\newblock


\bibitem[Mao et~al\mbox{.}(2021b)]%
        {seu}
\bibfield{author}{\bibinfo{person}{Xin Mao}, \bibinfo{person}{Wenting Wang}, \bibinfo{person}{Yuanbin Wu}, {and} \bibinfo{person}{Man Lan}.} \bibinfo{year}{2021}\natexlab{b}.
\newblock \showarticletitle{From Alignment to Assignment: Frustratingly Simple Unsupervised Entity Alignment}. In \bibinfo{booktitle}{\emph{Proceedings of the 2021 Conference on Empirical Methods in Natural Language Processing}}. \bibinfo{publisher}{Association for Computational Linguistics}, \bibinfo{address}{Online and Punta Cana, Dominican Republic}, \bibinfo{pages}{2843--2853}.
\newblock
\urldef\tempurl%
\url{https://doi.org/10.18653/v1/2021.emnlp-main.226}
\showDOI{\tempurl}


\bibitem[Mao et~al\mbox{.}(2022)]%
        {lightea}
\bibfield{author}{\bibinfo{person}{Xin Mao}, \bibinfo{person}{Wenting Wang}, \bibinfo{person}{Yuanbin Wu}, {and} \bibinfo{person}{Man Lan}.} \bibinfo{year}{2022}\natexlab{}.
\newblock \showarticletitle{{L}ight{EA}: A Scalable, Robust, and Interpretable Entity Alignment Framework via Three-view Label Propagation}. In \bibinfo{booktitle}{\emph{Proceedings of the 2022 Conference on Empirical Methods in Natural Language Processing}}, \bibfield{editor}{\bibinfo{person}{Yoav Goldberg}, \bibinfo{person}{Zornitsa Kozareva}, {and} \bibinfo{person}{Yue Zhang}} (Eds.). \bibinfo{publisher}{Association for Computational Linguistics}, \bibinfo{address}{Abu Dhabi, United Arab Emirates}, \bibinfo{pages}{825--838}.
\newblock
\urldef\tempurl%
\url{https://doi.org/10.18653/v1/2022.emnlp-main.52}
\showDOI{\tempurl}


\bibitem[Mena et~al\mbox{.}(2018)]%
        {sinkhorn2}
\bibfield{author}{\bibinfo{person}{Gonzalo Mena}, \bibinfo{person}{David Belanger}, \bibinfo{person}{Scott Linderman}, {and} \bibinfo{person}{Jasper Snoek}.} \bibinfo{year}{2018}\natexlab{}.
\newblock \showarticletitle{Learning Latent Permutations with Gumbel-Sinkhorn Networks}. In \bibinfo{booktitle}{\emph{International Conference on Learning Representations}}.
\newblock
\urldef\tempurl%
\url{https://openreview.net/forum?id=Byt3oJ-0W}
\showURL{%
\tempurl}


\bibitem[Pennington et~al\mbox{.}(2014)]%
        {glove}
\bibfield{author}{\bibinfo{person}{Jeffrey Pennington}, \bibinfo{person}{Richard Socher}, {and} \bibinfo{person}{Christopher Manning}.} \bibinfo{year}{2014}\natexlab{}.
\newblock \showarticletitle{{G}lo{V}e: Global Vectors for Word Representation}. In \bibinfo{booktitle}{\emph{Proceedings of the 2014 Conference on Empirical Methods in Natural Language Processing ({EMNLP})}}. \bibinfo{publisher}{Association for Computational Linguistics}, \bibinfo{address}{Doha, Qatar}, \bibinfo{pages}{1532--1543}.
\newblock
\urldef\tempurl%
\url{https://doi.org/10.3115/v1/D14-1162}
\showDOI{\tempurl}


\bibitem[Ribeiro et~al\mbox{.}(2021)]%
        {graph2text}
\bibfield{author}{\bibinfo{person}{Leonardo~FR Ribeiro}, \bibinfo{person}{Martin Schmitt}, \bibinfo{person}{Hinrich Sch{\"u}tze}, {and} \bibinfo{person}{Iryna Gurevych}.} \bibinfo{year}{2021}\natexlab{}.
\newblock \showarticletitle{Investigating Pretrained Language Models for Graph-to-Text Generation}. In \bibinfo{booktitle}{\emph{Proceedings of the 3rd Workshop on Natural Language Processing for Conversational AI}}. \bibinfo{pages}{211--227}.
\newblock


\bibitem[Suchanek et~al\mbox{.}(2007)]%
        {kgs_yago}
\bibfield{author}{\bibinfo{person}{Fabian~M. Suchanek}, \bibinfo{person}{Gjergji Kasneci}, {and} \bibinfo{person}{Gerhard Weikum}.} \bibinfo{year}{2007}\natexlab{}.
\newblock \showarticletitle{Yago: A Core of Semantic Knowledge}. In \bibinfo{booktitle}{\emph{Proceedings of the 16th International Conference on World Wide Web}} (Banff, Alberta, Canada) \emph{(\bibinfo{series}{WWW '07})}. \bibinfo{publisher}{Association for Computing Machinery}, \bibinfo{address}{New York, NY, USA}, \bibinfo{pages}{697–706}.
\newblock
\showISBNx{9781595936547}
\urldef\tempurl%
\url{https://doi.org/10.1145/1242572.1242667}
\showDOI{\tempurl}


\bibitem[Sun et~al\mbox{.}(2017)]%
        {dataset_dbp15k}
\bibfield{author}{\bibinfo{person}{Zequn Sun}, \bibinfo{person}{Wei Hu}, {and} \bibinfo{person}{Chengkai Li}.} \bibinfo{year}{2017}\natexlab{}.
\newblock \showarticletitle{Cross-Lingual Entity Alignment via Joint Attribute-Preserving Embedding}. In \bibinfo{booktitle}{\emph{The Semantic Web -- ISWC 2017}}, \bibfield{editor}{\bibinfo{person}{Claudia d'Amato}, \bibinfo{person}{Miriam Fernandez}, \bibinfo{person}{Valentina Tamma}, \bibinfo{person}{Freddy Lecue}, \bibinfo{person}{Philippe Cudr{\'e}-Mauroux}, \bibinfo{person}{Juan Sequeda}, \bibinfo{person}{Christoph Lange}, {and} \bibinfo{person}{Jeff Heflin}} (Eds.). \bibinfo{publisher}{Springer International Publishing}, \bibinfo{address}{Cham}, \bibinfo{pages}{628--644}.
\newblock
\showISBNx{978-3-319-68288-4}


\bibitem[Sun et~al\mbox{.}(2018)]%
        {bootea}
\bibfield{author}{\bibinfo{person}{Zequn Sun}, \bibinfo{person}{Wei Hu}, \bibinfo{person}{Qingheng Zhang}, {and} \bibinfo{person}{Yuzhong Qu}.} \bibinfo{year}{2018}\natexlab{}.
\newblock \showarticletitle{Bootstrapping Entity Alignment with Knowledge Graph Embedding}. In \bibinfo{booktitle}{\emph{Proceedings of the 27th International Joint Conference on Artificial Intelligence}} (Stockholm, Sweden) \emph{(\bibinfo{series}{IJCAI'18})}. \bibinfo{publisher}{AAAI Press}, \bibinfo{pages}{4396–4402}.
\newblock
\showISBNx{9780999241127}


\bibitem[Tang et~al\mbox{.}(2023b)]%
        {fgwea}
\bibfield{author}{\bibinfo{person}{Jianheng Tang}, \bibinfo{person}{Kangfei Zhao}, {and} \bibinfo{person}{Jia Li}.} \bibinfo{year}{2023}\natexlab{b}.
\newblock \showarticletitle{A Fused Gromov-Wasserstein Framework for Unsupervised Knowledge Graph Entity Alignment}. In \bibinfo{booktitle}{\emph{Annual Meeting of the Association for Computational Linguistics}}.
\newblock


\bibitem[Tang et~al\mbox{.}(2023a)]%
        {peea}
\bibfield{author}{\bibinfo{person}{Wei Tang}, \bibinfo{person}{Fenglong Su}, \bibinfo{person}{Haifeng Sun}, \bibinfo{person}{Q. Qi}, \bibinfo{person}{Jingyu Wang}, \bibinfo{person}{Shimin Tao}, {and} \bibinfo{person}{Hao Yang}.} \bibinfo{year}{2023}\natexlab{a}.
\newblock \showarticletitle{Weakly Supervised Entity Alignment with Positional Inspiration}.
\newblock \bibinfo{journal}{\emph{Proceedings of the Sixteenth ACM International Conference on Web Search and Data Mining}} (\bibinfo{year}{2023}).
\newblock


\bibitem[Tang et~al\mbox{.}(2020)]%
        {BERT-INT}
\bibfield{author}{\bibinfo{person}{Xiaobin Tang}, \bibinfo{person}{Jing Zhang}, \bibinfo{person}{Bo Chen}, \bibinfo{person}{Yang Yang}, \bibinfo{person}{Hong Chen}, {and} \bibinfo{person}{Cuiping Li}.} \bibinfo{year}{2020}\natexlab{}.
\newblock \showarticletitle{BERT-INT: a BERT-based interaction model for knowledge graph alignment}.
\newblock \bibinfo{journal}{\emph{interactions}}  \bibinfo{volume}{100} (\bibinfo{year}{2020}), \bibinfo{pages}{e1}.
\newblock


\bibitem[Wang et~al\mbox{.}(2019)]%
        {rec1}
\bibfield{author}{\bibinfo{person}{Hongwei Wang}, \bibinfo{person}{Fuzheng Zhang}, \bibinfo{person}{Miao Zhao}, \bibinfo{person}{Wenjie Li}, \bibinfo{person}{Xing Xie}, {and} \bibinfo{person}{Minyi Guo}.} \bibinfo{year}{2019}\natexlab{}.
\newblock \showarticletitle{Multi-Task Feature Learning for Knowledge Graph Enhanced Recommendation}. In \bibinfo{booktitle}{\emph{The World Wide Web Conference}} (San Francisco, CA, USA) \emph{(\bibinfo{series}{WWW '19})}. \bibinfo{publisher}{Association for Computing Machinery}, \bibinfo{address}{New York, NY, USA}, \bibinfo{pages}{2000–2010}.
\newblock
\showISBNx{9781450366748}
\urldef\tempurl%
\url{https://doi.org/10.1145/3308558.3313411}
\showDOI{\tempurl}


\bibitem[Wang et~al\mbox{.}(2020a)]%
        {minilm}
\bibfield{author}{\bibinfo{person}{Wenhui Wang}, \bibinfo{person}{Furu Wei}, \bibinfo{person}{Li Dong}, \bibinfo{person}{Hangbo Bao}, \bibinfo{person}{Nan Yang}, {and} \bibinfo{person}{Ming Zhou}.} \bibinfo{year}{2020}\natexlab{a}.
\newblock \showarticletitle{Minilm: Deep self-attention distillation for task-agnostic compression of pre-trained transformers}.
\newblock \bibinfo{journal}{\emph{Advances in Neural Information Processing Systems}}  \bibinfo{volume}{33} (\bibinfo{year}{2020}), \bibinfo{pages}{5776--5788}.
\newblock


\bibitem[Wang et~al\mbox{.}(2022)]%
        {rec3}
\bibfield{author}{\bibinfo{person}{Xiaolei Wang}, \bibinfo{person}{Kun Zhou}, \bibinfo{person}{Ji-Rong Wen}, {and} \bibinfo{person}{Wayne~Xin Zhao}.} \bibinfo{year}{2022}\natexlab{}.
\newblock \showarticletitle{Towards unified conversational recommender systems via knowledge-enhanced prompt learning}. In \bibinfo{booktitle}{\emph{Proceedings of the 28th ACM SIGKDD Conference on Knowledge Discovery and Data Mining}}. \bibinfo{pages}{1929--1937}.
\newblock


\bibitem[Wang et~al\mbox{.}(2018)]%
        {gcn-align}
\bibfield{author}{\bibinfo{person}{Zhichun Wang}, \bibinfo{person}{Qingsong Lv}, \bibinfo{person}{Xiaohan Lan}, {and} \bibinfo{person}{Yu Zhang}.} \bibinfo{year}{2018}\natexlab{}.
\newblock \showarticletitle{Cross-lingual Knowledge Graph Alignment via Graph Convolutional Networks}. In \bibinfo{booktitle}{\emph{Proceedings of the 2018 Conference on Empirical Methods in Natural Language Processing}}. \bibinfo{publisher}{Association for Computational Linguistics}, \bibinfo{address}{Brussels, Belgium}, \bibinfo{pages}{349--357}.
\newblock
\urldef\tempurl%
\url{https://doi.org/10.18653/v1/D18-1032}
\showDOI{\tempurl}


\bibitem[Wang et~al\mbox{.}(2020b)]%
        {EPEA}
\bibfield{author}{\bibinfo{person}{Zhichun Wang}, \bibinfo{person}{Jinjian Yang}, {and} \bibinfo{person}{Xiaoju Ye}.} \bibinfo{year}{2020}\natexlab{b}.
\newblock \showarticletitle{Knowledge Graph Alignment with Entity-Pair Embedding}. In \bibinfo{booktitle}{\emph{Proceedings of the 2020 Conference on Empirical Methods in Natural Language Processing (EMNLP)}}. \bibinfo{publisher}{Association for Computational Linguistics}, \bibinfo{address}{Online}, \bibinfo{pages}{1672--1680}.
\newblock
\urldef\tempurl%
\url{https://doi.org/10.18653/v1/2020.emnlp-main.130}
\showDOI{\tempurl}


\bibitem[Wu et~al\mbox{.}(2019b)]%
        {rdgcn}
\bibfield{author}{\bibinfo{person}{Yuting Wu}, \bibinfo{person}{Xiao Liu}, \bibinfo{person}{Yansong Feng}, \bibinfo{person}{Zheng Wang}, \bibinfo{person}{Rui Yan}, {and} \bibinfo{person}{Dongyan Zhao}.} \bibinfo{year}{2019}\natexlab{b}.
\newblock \showarticletitle{Relation-Aware Entity Alignment for Heterogeneous Knowledge Graphs}. In \bibinfo{booktitle}{\emph{Proceedings of the Twenty-Eighth International Joint Conference on Artificial Intelligence, {IJCAI-19}}}. \bibinfo{publisher}{International Joint Conferences on Artificial Intelligence Organization}, \bibinfo{pages}{5278--5284}.
\newblock
\urldef\tempurl%
\url{https://doi.org/10.24963/ijcai.2019/733}
\showDOI{\tempurl}


\bibitem[Wu et~al\mbox{.}(2019a)]%
        {hgcn}
\bibfield{author}{\bibinfo{person}{Yuting Wu}, \bibinfo{person}{Xiao Liu}, \bibinfo{person}{Yansong Feng}, \bibinfo{person}{Zheng Wang}, {and} \bibinfo{person}{Dongyan Zhao}.} \bibinfo{year}{2019}\natexlab{a}.
\newblock \showarticletitle{Jointly Learning Entity and Relation Representations for Entity Alignment}. In \bibinfo{booktitle}{\emph{Conference on Empirical Methods in Natural Language Processing}}.
\newblock
\urldef\tempurl%
\url{https://api.semanticscholar.org/CorpusID:202712648}
\showURL{%
\tempurl}


\bibitem[Xu et~al\mbox{.}(2019a)]%
        {jape}
\bibfield{author}{\bibinfo{person}{Kun Xu}, \bibinfo{person}{Liwei Wang}, \bibinfo{person}{Mo Yu}, \bibinfo{person}{Yansong Feng}, \bibinfo{person}{Yan Song}, \bibinfo{person}{Zhiguo Wang}, {and} \bibinfo{person}{Dong Yu}.} \bibinfo{year}{2019}\natexlab{a}.
\newblock \showarticletitle{Cross-lingual Knowledge Graph Alignment via Graph Matching Neural Network}. In \bibinfo{booktitle}{\emph{Proceedings of the 57th Annual Meeting of the Association for Computational Linguistics}}. \bibinfo{publisher}{Association for Computational Linguistics}, \bibinfo{address}{Florence, Italy}, \bibinfo{pages}{3156--3161}.
\newblock
\urldef\tempurl%
\url{https://doi.org/10.18653/v1/P19-1304}
\showDOI{\tempurl}


\bibitem[Xu et~al\mbox{.}(2019b)]%
        {gm-align}
\bibfield{author}{\bibinfo{person}{Kun Xu}, \bibinfo{person}{Liwei Wang}, \bibinfo{person}{Mo Yu}, \bibinfo{person}{Yansong Feng}, \bibinfo{person}{Yan Song}, \bibinfo{person}{Zhiguo Wang}, {and} \bibinfo{person}{Dong Yu}.} \bibinfo{year}{2019}\natexlab{b}.
\newblock \showarticletitle{Cross-lingual Knowledge Graph Alignment via Graph Matching Neural Network}. In \bibinfo{booktitle}{\emph{Proceedings of the 57th Annual Meeting of the Association for Computational Linguistics}}. \bibinfo{publisher}{Association for Computational Linguistics}, \bibinfo{address}{Florence, Italy}, \bibinfo{pages}{3156--3161}.
\newblock
\urldef\tempurl%
\url{https://doi.org/10.18653/v1/P19-1304}
\showDOI{\tempurl}


\bibitem[Zeng et~al\mbox{.}(2020)]%
        {CEA}
\bibfield{author}{\bibinfo{person}{Weixin Zeng}, \bibinfo{person}{Xiang Zhao}, \bibinfo{person}{Jiuyang Tang}, {and} \bibinfo{person}{Xuemin Lin}.} \bibinfo{year}{2020}\natexlab{}.
\newblock \showarticletitle{Collective Entity Alignment via Adaptive Features}. In \bibinfo{booktitle}{\emph{2020 IEEE 36th International Conference on Data Engineering (ICDE)}}. \bibinfo{pages}{1870--1873}.
\newblock
\urldef\tempurl%
\url{https://doi.org/10.1109/ICDE48307.2020.00191}
\showDOI{\tempurl}


\bibitem[Zhang et~al\mbox{.}(2024)]%
        {srea}
\bibfield{author}{\bibinfo{person}{Yuhong Zhang}, \bibinfo{person}{Jianqing Wu}, \bibinfo{person}{Kui Yu}, {and} \bibinfo{person}{Xindong Wu}.} \bibinfo{year}{2024}\natexlab{}.
\newblock \showarticletitle{Diverse Structure-Aware Relation Representation in Cross-Lingual Entity Alignment}.
\newblock \bibinfo{journal}{\emph{ACM Trans. Knowl. Discov. Data}} \bibinfo{volume}{18}, \bibinfo{number}{4}, Article \bibinfo{articleno}{92} (\bibinfo{date}{Feb.} \bibinfo{year}{2024}), \bibinfo{numpages}{23}~pages.
\newblock
\showISSN{1556-4681}
\urldef\tempurl%
\url{https://doi.org/10.1145/3638778}
\showDOI{\tempurl}


\bibitem[Zhao et~al\mbox{.}(2020)]%
        {qa1}
\bibfield{author}{\bibinfo{person}{Chen Zhao}, \bibinfo{person}{Chenyan Xiong}, \bibinfo{person}{Xin Qian}, {and} \bibinfo{person}{Jordan Boyd-Graber}.} \bibinfo{year}{2020}\natexlab{}.
\newblock \showarticletitle{Complex Factoid Question Answering with a Free-Text Knowledge Graph}. In \bibinfo{booktitle}{\emph{Proceedings of The Web Conference 2020}} (Taipei, Taiwan) \emph{(\bibinfo{series}{WWW '20})}. \bibinfo{publisher}{Association for Computing Machinery}, \bibinfo{address}{New York, NY, USA}, \bibinfo{pages}{1205–1216}.
\newblock
\showISBNx{9781450370233}
\urldef\tempurl%
\url{https://doi.org/10.1145/3366423.3380197}
\showDOI{\tempurl}


\bibitem[Zhu et~al\mbox{.}(2022)]%
        {RAEA}
\bibfield{author}{\bibinfo{person}{Beibei Zhu}, \bibinfo{person}{Tie Bao}, \bibinfo{person}{Lu Liu}, \bibinfo{person}{Jiayu Han}, \bibinfo{person}{Junyi Wang}, {and} \bibinfo{person}{Tao Peng}.} \bibinfo{year}{2022}\natexlab{}.
\newblock \showarticletitle{Cross-lingual knowledge graph entity alignment based on relation awareness and attribute involvement}.
\newblock \bibinfo{journal}{\emph{Applied Intelligence}}  \bibinfo{volume}{53} (\bibinfo{year}{2022}), \bibinfo{pages}{6159--6177}.
\newblock
\urldef\tempurl%
\url{https://api.semanticscholar.org/CorpusID:250372887}
\showURL{%
\tempurl}


\bibitem[Zhu et~al\mbox{.}(2021)]%
        {raga}
\bibfield{author}{\bibinfo{person}{Renbo Zhu}, \bibinfo{person}{Meng Ma}, {and} \bibinfo{person}{Ping Wang}.} \bibinfo{year}{2021}\natexlab{}.
\newblock \showarticletitle{RAGA: Relation-Aware Graph Attention Networks for Global Entity Alignment}. In \bibinfo{booktitle}{\emph{Advances in Knowledge Discovery and Data Mining: 25th Pacific-Asia Conference, PAKDD 2021, Virtual Event, May 11–14, 2021, Proceedings, Part I}}. \bibinfo{publisher}{Springer-Verlag}, \bibinfo{address}{Berlin, Heidelberg}, \bibinfo{pages}{501–513}.
\newblock
\showISBNx{978-3-030-75761-8}
\urldef\tempurl%
\url{https://doi.org/10.1007/978-3-030-75762-5_40}
\showDOI{\tempurl}


\end{thebibliography}
